%% file: main.tex
\documentclass{article} %
\usepackage[utf8]{inputenc}
\usepackage[T1]{fontenc}

\usepackage{style}
\input{math_commands.tex}

\usepackage{wrapfig}
\usepackage{graphicx}
\usepackage{adjustbox}
\usepackage{booktabs}
\usepackage{amssymb}
\usepackage{enumitem}
\usepackage{multicol}

\usepackage{pifont}

\usepackage{xspace}
\usepackage{booktabs}
\usepackage{graphicx}
\usepackage[newfloat=true,frozencache,cachedir=.]{minted}
\usepackage{mdframed}
\usepackage{fancyvrb}
\usepackage{cleveref}
\usepackage{nicematrix}
\usepackage{authblk}
\usepackage{nicefrac}

\newcommand{\bint}{\texttt{int8}}
\newcommand{\indexint}{\texttt{IndexInt}}

\title{\sc What Algorithms can Transformers Learn?\\
A Study in Length Generalization}

\author[1,2]{Hattie Zhou\thanks{Work done while interning at Apple.}}
\author[1]{Arwen Bradley}
\author[1]{Etai Littwin}
\author[1,3]{Noam Razin\protect\footnotemark[1]}
\author[1]{Omid Saremi}
\author[1]{\authorcr Josh Susskind}
\author[1]{Samy Bengio}
\author[1]{Preetum Nakkiran}
\affil[1]{Apple}
\affil[2]{Mila, Université de Montréal}
\affil[3]{Tel Aviv University}

\date{}

\newcommand{\Vocab}{\mathcal{V}}

\begin{document}

\maketitle

\begin{abstract}

Large language models exhibit surprising emergent generalization properties,
yet also struggle on many simple reasoning tasks such as arithmetic and parity.
This raises the question of if and when Transformer models
can learn the true algorithm for solving a task.
We study the scope of Transformers' abilities
in the specific setting of length generalization on algorithmic tasks.
Here, we propose a unifying framework
to understand when and how Transformers can
exhibit strong length generalization on a given task.
Specifically, we leverage RASP \citep{weiss2021thinking}--- a programming language
designed for the computational model of a Transformer---
and introduce the RASP-Generalization Conjecture:
Transformers tend to length generalize
on a task if the task can be solved by a short RASP program
which works for all input lengths.
This simple conjecture remarkably captures most known
instances of length generalization on algorithmic tasks. Moreover, we leverage our insights to drastically improve generalization performance on traditionally hard tasks (such as parity and addition).
On the theoretical side,
we give a simple example where the ``min-degree-interpolator'' model of learning
from \citet{abbe2023generalization} does not correctly
predict Transformers' out-of-distribution behavior, but our conjecture does.
Overall, our work provides a novel perspective on the
mechanisms of compositional generalization and the algorithmic capabilities of Transformers.
\end{abstract}

\input{intro2}

\input{conjecture}
\input{prelim}

\input{rasp}

\input{scratchpads}

\input{gotu}

\input{discussion}

\section{Conclusion}

There has been significant interest recently in understanding and
improving the reasoning abilities of Transformer-based models \citep{cobbe2021training, magister2023teaching, wei2022chain, FF, suzgun2022challenging, lewkowycz2022solving, press2023measuring}.
An important step towards this goal is to understand the conditions
under which a Transformer model can learn general problem-solving strategies
or algorithms simply by training on examples--- if at all possible.
We study this question in a controlled setting by focusing on length generalization
as an instance of algorithm learning, with standard decoder-only Transformers trained from scratch on synthetic algorithmic tasks.

The guiding philosophy behind our work is that we should
think of Transformers not
not as \emph{functions} with a fixed input size,
but as \emph{programs}, defined for inputs of all lengths.
With this perspective,
we conjecture that
algorithms which are simple-to-represent by a Transformer
are also more likely to be learned. 
To operationalize this, we 
constructed the RASP-L language based on \citet{weiss2021thinking},
and then defined ``simple-to-represent'' 
as ``can be written as a short RASP-L program.''
Remarkably, our RASP conjecture captures most if
not all known instances of length generalization,
and provides a unifying perspective on when length generalization
and algorithm learning are possible.
The tools developed here could in theory apply to 
compositional generalization more broadly, beyond
length generalization; we consider
this a fruitful direction for future work.
Overall, we hope our results help demystify certain
observations of the surprising ``reasoning-like'' abilities of
Transformers, by showing that some of these behaviors
can actually emerge for simple, unsurprising reasons.

\subsubsection*{Acknowledgements}

We thank (alphabetically) 
Samira Abnar,
Madhu Advani,
Jaros\l aw B\l asiok,
Stefano	Cosentino,
Laurent Dinh,
Fartash Faghri,
Spencer Frei,
Yejin Huh,
Vaishaal Shankar,
Vimal Thilak,
Russ Webb,
Jason Yosinski,
and
Fred Zhang
for feedback on early drafts and discussions throughout the project.

\bibliography{iclr2024_conference}
\bibliographystyle{iclr2024_conference}
\newpage
\appendix
\input{related}
\input{appendix}

\end{document}

%% file: math_commands.tex
\usepackage{amsmath,amsfonts,bm,amsthm}

\newtheorem{theorem}{Theorem}
\newtheorem{lemma}[theorem]{Lemma}

\renewcommand{\bar}{\overline}
\renewcommand{\epsilon}{\varepsilon}

\def\eqref#1{equation~\ref{#1}}

\def\1{\bm{1}}

\DeclareMathAlphabet{\mathsfit}{\encodingdefault}{\sfdefault}{m}{sl}
\SetMathAlphabet{\mathsfit}{bold}{\encodingdefault}{\sfdefault}{bx}{n}

\DeclareMathOperator*{\argmax}{arg\,max}
\DeclareMathOperator*{\argmin}{arg\,min}

%% file: intro2.tex
\section{Introduction}

Large language models (LLMs) have shown impressive abilities in natural language generation, reading comprehension, code-synthesis, instruction-following, commonsense reasoning, and many other tasks \citep{brown2020language, chen2021evaluating, chowdhery2022palm, minerva, gunasekar2023textbooks, touvron2023llama}.
However, when evaluated in controlled studies, Transformers often struggle with out-of-distribution generalization \citep{nogueira2021investigating,ontañón2022making, FF, wu2023reasoning, saparov2023testing}.
It is thus not clear how to reconcile
Transformers' seemingly-impressive performance in some settings
with their fragility in others.

In this work, we aim to understand when standard decoder-only Transformers can
generalize systematically beyond their training distribution.
We adopt the approach of recent studies and focus on length generalization on algorithmic tasks as a measure of how well language models can learn to reason \citep{nogueira2021investigating, kim-etal-2021-seen, anil2022exploring, lee2023teaching, FF, welleck2022symbolic, liu2023transformers2}. 
Length generalization evaluates the model on problems that are longer (and harder) than seen in the training set--- for example, training on 10 digit decimal addition problems, and testing on 20 digit problems. This challenging evaluation setting gives an indication of whether the model has learned the correct underlying algorithm for a given task. 

There is currently no clear answer in the literature about when (or even if) Transformers length generalize.
Transformers trained from scratch on addition and other arithmetic tasks exhibit little to no
length generalization in prior work \citep{nye2021show, nogueira2021investigating, lee2023teaching}, and even models finetuned from pretrained LLMs struggle on simple algorithmic tasks \citep{anil2022exploring}.
Going even further, \citet{FF} hypothesize that Transformers solve tasks via ``analogical pattern matching'',
and thus fundamentally cannot acquire systematic problem-solving capabilities.
On the other hand, length generalization can sometimes occur for 
particular architectural choices and scratchpad formats \citep{jelassi2023length,kazemnejad2023impact, li2023representations}.
A number of papers have studied this question for specific classes of problems, such as
decimal addition \citep{lee2023teaching},
Dyck-1 languages \citep{bhattamishra2020ability},
Boolean functions \citep{abbe2023generalization},
structural recursion \citep{zhang2023transformers},
and finite state automaton \citep{liu2023transformers2}.
However, there is no unifying framework for reasoning about length generalization in Transformers
which captures both their surprising failures \emph{and} surprising capabilities, and applies
to a broad class of symbolic reasoning tasks.

As a starting point of our work, we show that there exist algorithmic tasks where
Transformers trained from scratch generalize naturally far outside of the training distribution.
This observation contradicts the conventional wisdom in much of the existing literature,
and suggests that length generalization is
not inherently problematic for the Transformer architecture--- though it clearly does not occur for all tasks.
\emph{Why then do Transformers exhibit strong length generalization on
certain tasks and not others, and what are the mechanisms behind generalization when it occurs?}
In the following sections, we will propose a unifying framework to predict cases
of successful length generalization and describe a possible underlying mechanism.

\paragraph{Preview of Length Generalization.}

\begin{figure}[t]
\captionsetup[subfigure]{justification=centering}
	\centering
    \subfloat[Transformer length generalization]{
\begin{adjustbox}{max width=8cm}
\begin{tabular}{lp{1.8cm}lp{1.8cm}}\toprule 
\multicolumn{2}{c}{\textbf{RASP-L Program Exists}}  & \multicolumn{2}{c}{\textbf{No RASP-L Program}} \\
\cmidrule(lr){1-2}\cmidrule(lr){3-4}
Task Name & $\textrm{Test EM}$ (+10 length)   & Task Name & $\textrm{Test EM}$ (+10 length) \\
\midrule
Count & 100\%  & &  \\
Mode & 100\%  & Mode (hard) & 0\%   \\
Unique Copy & 96\% & Repeat Copy & 0\% \\
Sort & 96\%  &  &  \\
Addition (new) & 100\%   & Addition & 0\%  \\
Parity (new) & 100\%   & Parity & 0\%   \\
\bottomrule
\end{tabular}
\end{adjustbox}
\label{fig:hero}}
	\subfloat[Length generalization on counting task]{\includegraphics[width=.4\linewidth]{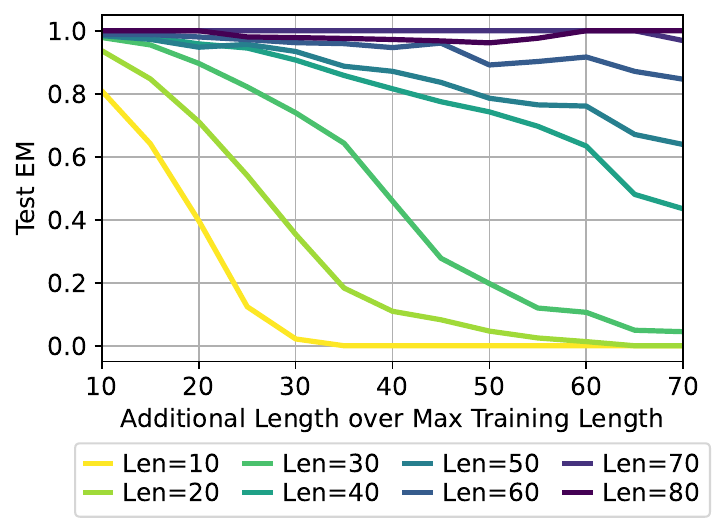}\label{fig:count-lengen-scale}} 
	\caption{{\bf \protect\subref{fig:hero}}
 A selection of tasks studied in this paper partitioned by whether
 they can be solved by programs in the RASP-L programming language (discussed in Section~\ref{sec:rasp}).
 Test EM denotes the exact match accuracy (EM) for test inputs of length 10 greater than train.
 We show that all tasks which admit a short solution in the RASP-L programming language
 also exhibit strong length generalization performance, and vice versa. 
 For certain hard tasks we construct ``new'' versions 
 which admit RASP-L solutions,
 by carefully modifying the input and scratchpad format, and these versions
 length generalize.
 We also show how poor scratchpad formats can make tasks harder, by giving an example for the Mode task.
    {\bf \protect\subref{fig:count-lengen-scale}} Length generalization
    for the counting task (described below), measured by exact match accuracy (EM).
    Transformers are trained on sequences of varying length,
    and tested at different levels of out-of-distribution
    over the maximum training length.
    Models trained on sequences of length $60$ or more exhibit near perfect length generalization
    up to length $150$ (max evaluation length).
    }
\label{fig:count-lengen}
\end{figure}

We begin by introducing a simple task that exhibits strong length generalization,
as a minimal example of the phenomenon.
The task is ``counting'': given a prompt%
\texttt{
\setlength{\tabcolsep}{3pt}
\small
\begin{NiceTabular}{*{4}{c}}[hvlines]
SoS & a & b & >
\end{NiceTabular}
}
for numbers $a, b$,
the model must count from $a$ to $b$ inclusive, and terminate with ``\verb|EoS|''.
An example is:
\texttt{
\setlength{\tabcolsep}{3pt}
\small
\begin{NiceTabular}{*{9}{c}}[hvlines]
SoS & 2 & 5 & > & 2 & 3 & 4 & 5 & EoS
\end{NiceTabular}
}.
We train a Transformer with learned positional embeddings
on count sequences of lengths up to $50$, with random endpoints
$a$ and $b$ in $[0, 155]$.
We ``pack the context'' with i.i.d. sequences during training,
following standard practices in LLM pipelines.
This trained model then length-generalizes near perfectly when prompted to count sequences of length $100$ (see Figure~\ref{fig:count-lengen-scale}).

\paragraph{Possible Mechanisms.}

To understand the above example, it is helpful to first consider:
why should length generalization be possible at all?
The crucial observation is that the Transformer
architecture is already equipped with a natural notion of length-extension.
If we omit positional encodings for simplicity, 
then a fixed setting of Transformer weights defines a sequence-to-sequence function on 
sequences of \emph{arbitrary length}.
If this function applies the correct transformation
for inputs of any length in the context, then we can expect it to length generalize.

For the count task, length generalization is possible if the
model somehow learns a correct algorithm to solve the count task. One such algorithm is as follows. To predict the next token:

\hspace*{0.2cm}%
\begin{minipage}{.95\textwidth}
\begin{enumerate}
    \item Search for the most-recent \Verb|SoS| token, 
    and read the following two numbers as $a, b$.
    \item Read the latest output token as $x$.
    If \verb|(x=='>')|, output $a$.
    If \verb|(x==b)|, output \Verb|EoS|.
    \item Otherwise, output $(x+1)$.
\end{enumerate}
\end{minipage}

This program applies to sequences of all lengths. Thus, if the model ends up learning the algorithm from short sequences,
then it will automatically length-generalize to long sequences.
The discussion so far could apply to any auto-regressive model, not just Transformers.
What is special about the Transformer architecture and the count task, though, is that a Transformer can \emph{easily represent}
the above program, uniformly for all input lengths.
This is not trivial:
we claim that the same exact Transformer weights
which solve the task at length 20, can also solve the task at length 50 and greater lengths (Figure~\ref{fig:hero}). Interestingly, we find supporting evidence in Appendix~\ref{app:counerfactual} that the trained models are actually implementing this algorithm.

\paragraph{Overview.}
The core message of this work is that it is actually possible for
Transformers to approximately
learn a length-generalizing algorithm,
if the correct algorithm is both possible and ``simple'' 
to represent with a Transformer.
In our main conjecture in Section~\ref{sec:conjecture},
we propose a set of conditions that determine whether a Transformer model
trained from scratch is likely to generalize on a given symbolic task.
Our conjecture builds upon the following intuition:

\begin{framed}
{\bf Intuition.}
\textit{
Algorithms which are simple to represent by Transformers are also easy to learn for Transformers, and vice versa.
}
\end{framed}

This intuition tells us that to reason about whether Transformers will length-generalize on a given task, we should consider whether the underlying algorithm that solves the task is naturally representable by the Transformer architecture. To do so, we leverage the recently introduced RASP programming language \citep{weiss2021thinking, lindner2023tracr},
which is essentially an ``assembly language for Transformers.''
We describe RASP, and our variant of it RASP-L, in Section~\ref{sec:rasp}.
For now, it suffices to say RASP-L is a human-readable programming language
which defines 
programs that can be compiled into Transformer weights\footnote{In fact \citet{lindner2023tracr} provide an explicit compiler from RASP programs to Transformer weights.},
such that each line of RASP-L compiles into at most one Transformer layer.
RASP-L lets us reason about Transformer-representable algorithms at 
a familiar level of abstraction--- similar to standard programming languages like Python,
but for programs which ``run on a Transformer'' instead of running on a standard von Neumann computer.
We leverage the RASP-L language to state a more precise version of our intuition,
which predicts exactly which functions Transformers learn:

\begin{framed}
{\bf Toy Model (RASP MDL Principle).}
\textit{
For symbolic tasks,
Transformers tend to learn the shortest RASP-L program which fits the training set.
Thus, if this minimal program correctly length-generalizes, then so will the Transformer.
}
\end{framed}

This toy model is similar to
Minimum-Description-Length principles (e.g. \citet{shalev2014understanding})
and Solomonoff induction \citep{solomonoff1964formal}.
The key insight is that we propose using a measure of complexity
that specifically corresponds to the unique information-flow inside Transformer architectures. 
Although many prior works have conjectured similar ``simplicity bias'' in Transformers \citep{abbe2023generalization, bhattamishra2023simplicity},
the notion of simplicity we use here is tailor-made for the Transformer architecture.
This distinction is important: In Section~\ref{sec:gotu},
we show a simple setting where the popular ``minimum polymomial degree'' notion of simplicity
\citep{abbe2023generalization}
does not correctly predict Transformer generalization, whereas our RASP-based notion of simplicity does.
Finally, notice that we use a notion of simplicity over \emph{programs},
rather than over functions with fixed input dimension,
which makes it especially suitable for understanding algorithmic tasks.

\subsection{Organization and Contributions}

In Section~\ref{sec:conjecture}, we introduce the RASP-Generalization Conjecture (RASP conjecture for short),
which uses the RASP-L programming language to predict
when Transformers are likely to length-generalize.
To help users develop intuition about RASP-L algorithms,
we present a ``standard library'' of RASP-L functions (Section~\ref{sec:rasp})
that can be used as modular components of programs.
This includes an implementation of ``induction heads'' \citep{olsson2022context}, for example.
In Section~\ref{sec:lengenexp}, we show that the RASP conjecture
is consistent with experiments on a variety of algorithmic tasks.
Then, in Section~\ref{sec:scratchpad}, we 
leverage our conjecture to improve length generalization, which results in the first instance of strong length generalization on parity and addition for Transformer models trained from scratch.
Finally, on the theoretical side, we give an example where
the ``min-degree-interpolator'' model of learning from \citet{abbe2023generalization}
does not produce correct predictions for Transformers,
but our conjecture does (Section~\ref{sec:gotu}).
We conclude with a discussion of limitations and open questions, 
and discuss additional related works in Appendix~\ref{app:relatedworks}.

%% file: conjecture.tex
\section{Main Conjecture}
\label{sec:conjecture}

We first set some notation.
A \emph{Transformer} \citep{vaswani2017attention}
refers to an instance of a
decoder-only causal Transformer architecture, with 
any fixed setting of weights, along with 
any computable positional embedding scheme\footnote{This is a technical detail:
we consider position encoding schemes which can be uniformly generated, i.e. 
there exists a Turing machine which on input $(i, n)$, produces the positional embedding
vector for index $i$ out of $n$ total.}.
As a technical point, we allow the transformer weights to take values in the
extended real line $\R \cup \{\pm \infty\}$, to allow saturating
the softmax at arbitrarily large context lengths\footnote{This bypasses the limitations presented in \citet{hahn2020theoretical}, which exploit non-saturating softmaxes.}.
We consider only greedy sampling throughout, since our tasks are deterministic.
In this conjecture, and throughout the paper,
we consider Transformers ``trained to completion,''
meaning trained to near-optimal performance on their training distribution.
That is, we assume that in-distribution generalization is achieved nearly-optimally,
and focus our attention on the induced out-of-distribution generalization.
The exact training procedure we consider is given in Section~\ref{sec:setup}.
We now describe our main conjecture.

{\bf RASP-Generalization Conjecture.}
{\itshape
A decoder-only autoregressive Transformer is likely to length-generalize when
trained to completion on an algorithmic task if the following conditions hold.
\begin{enumerate}
    \item \textbf{Realizability.} The true next-token function for the task 
    can be represented by a single causal Transformer %
    which works on all input lengths.
    \item \textbf{Simplicity.}
    This representation is ``simple'', meaning it can be written in
    RASP-L (a learnable subset of RASP defined in Section~\ref{sec:rasp}).
    \item \textbf{Diversity.}
    The training data is sufficiently diverse, such that there does not
    exist any shorter RASP-L program which agrees with
    the task in-distribution but not out-of-distribution.
\end{enumerate}
Remarkably, the above features are empirically correlated with:
(1) generalization to longer lengths out-of-distribution, and
(2) faster train optimization in-distribution.
}

The first condition of realizability is actually quite stringent,
because it requires a \emph{single Transformer} to be able to solve the task at \emph{all lengths}.
Causal Transformers define a particular computational model, 
and not all sequence-to-sequence tasks can be solved within this model.
For example, next-token functions which require $\Omega(n^3)$ computation time
on inputs of length $n$ provably cannot be represented by a Transformer, however large\footnote{%
Such tasks exist by the classical Time Hierarchy Theorem (e.g. \citet{arora2009computational})
since Transformers can be simulated by Turing machines in $O(n^2)$ time.}.
Now, the realizability condition may seem stronger than required,
because in practice, we do not actually need length generalization
for arbitrary unbounded lengths--- only lengths up to some maximum context size.
Nevertheless, we find that
considering representability in the unbounded length setting
is a good heuristic for learnability in bounded length settings.
Intuitively, if a task requires a different Transformer for each input length,
then it may be an ``unnatural'' task for Transformers,
and unlikely to generalize well.

We emphasize that our conjecture is primarily \emph{phenomenological},
as opposed to \emph{mechanistic}.
That is, we do not claim Transformers will
actually learn weights which are close to the compiled weights of the RASP-L program.
We only claim that RASP-L is a useful predictive tool:
empirically, if a task can be solved by a RASP-L program,
then it can be learned easily by a Transformer, and vice versa.
Although we believe the toy model is a plausible mechanism that implies
our conjecture, we leave investigating this more fully as an important question for future work.

%% file: prelim.tex
\subsection{Experimental Validation}
\label{sec:eval}

We empirically evaluate the conjecture by training Transformers on a set of algorithmic tasks in Section~\ref{sec:lengenexp}. The conjecture proposes three conditions: 1) realizability, 2) simplicity, and 3) diversity. Simplicity implies realizability, as being able to come up with a RASP-L program for a task guarantees realizability. In our experiments, we propose a set of ``easy'' tasks for which we could write simple RASP-L solutions for, and a set of ``hard'' tasks for which we could not, and evaluate the relationship between task difficulty (per RASP conjecture) and length generalization performance. 
Moreover, we evaluate the hypothesis that simple-to-represent programs are more easily learned by looking at the train convergence speed of tasks with varying RASP-L complexity. Lastly, although task diversity is a rich area for investigation, here we focus on varying the range of lengths seen during training as a simple proxy for diversity.

There are many other factors that may influence the generalization performance of Transformer models,
including architectural innovations, positional embeddings, and training methodology \citep{dehghani2019universal, furrer2021compositional, ontañón2022making, press2022train, ruoss-etal-2023-randomized}.
Such factors can also influence how robustly the correct solution is learned, if indeed it is.
Since our study focuses on the relevant characteristics of \emph{tasks},
and not of architectures, we employ standard decoder-only causal Transformers with learned positional embeddings throughout this paper.
If models trained in this simple setting can still exhibit non-trivial length generalization on a given task,
we can conclude that this task is amenable to generalization. For completeness, we also include results using rotary positional embedding in Appendix~\ref{app:rotary}. 

Note, we do not expect (and indeed do not observe) \emph{perfect} length generalization over arbitrary lengths. 
Some level of degradation is likely fundamental due to
issues of the noisy optimization process, continuous weight space,
finite-precision, etc.
Moreover, the model may learn a program that functionally approaches
the target program, but does not exactly match it.
Thus, we study the external phenomena of non-trivial length generalization performance in this paper,
and leave the mechanistic investigation of the learned programs to future work.

%% file: rasp.tex
\section{RASP-L: What Algorithms Can Transformers Learn?}
\label{sec:rasp}

We will now define a version of the RASP programming language \citep{weiss2021thinking},
which we call RASP-L.
RASP-L is essentially a restricted version of RASP, with one additional feature.
Since programming in RASP-L is fairly non-intuitive, we also introduce a ``standard library''
of useful RASP-L functions, which can be composed to solve more complex tasks.
We first review the original RASP language.

\subsection{Background: A RASP Primer (in Python)}

The original RASP language can be thought of as a domain-specific-language for specifying Transformer weights,
in human-readable form \citep{weiss2021thinking}.
Importantly, RASP was designed for the computational model of Transformers,
so short RASP programs define functions which are ``easy to represent'' for Transformers.
Although RASP was conceived as a separate language with its own syntax,
it is possible to realize RASP as
a restricted subset of Python where only a few operations are allowed.
We do this explicitly in Listing~\ref{rasp:core}, briefly described here\footnote{We plan to release
the Python code in these listings on Github.}.

Every RASP program accepts an input sequence of length $n$,
for all $n \in \N$, and
returns an output sequence of the exact same length--- just like a Transformer.
No control flow is allowed; all programs must be straight-line programs,
with no branching or loops.
Concretely, every line of a RASP program must be a call to one of the
core functions defined in Listing~\ref{rasp:core}, or to another RASP program.
The core operations allowed in RASP are: arbitrary elementwise operations over sequences
(\texttt{map} and \texttt{seq\_map}), and a very particular type of non-elementwise operation
\texttt{kqv}, which simulates a causal attention layer.
The \texttt{kqv} function takes as input three sequences and a binary predicate
(such as greater-than),
then constructs a boolean attention matrix by applying the predicate to all pairs of key
and query, and applies this matrix to the value sequence (see Listing~\ref{rasp:core}).

\paragraph{Causality.} Since we study autoregressive decoder-only Transformers,
we must use the \emph{causal} version of RASP, where all sequence-to-sequence operations
are executed causally.
Moreover, while RASP programs define sequence-to-sequence functions,
we interpret them as autoregressive functions
by taking the last token of output sequence
as the next-token prediction.
This setting differs from most prior literature on RASP, which typically consider non-causal models,
and these differences significantly change the nature of RASP programming.

\paragraph{Intuition.}
A key characteristic of RASP is that it only allows parallelizable operations,
because Transformers are an inherently \emph{parallel} model of computation.
This makes performing inherently-sequential computation, such as 
iterating through each input symbol and updating an internal state,
tricky if not impossible to write in RASP.
This is why loops are not allowed in RASP: a Transformer
has only constant depth, and cannot directly simulate 
an arbitrary number of loop iterations.
One way to bypass this limitation is to exploit
the \emph{autoregressive} inference procedure\footnote{A similar observation was presented in \citet{malach2023auto}.}.
Since the model is called iteratively at inference time,
this effectively provides an ``outer-loop''
that can enable a certain kind of sequential computation,
where the sequential state is encoded into the prior context.
This is exactly what scratchpads enable, as we elaborate in Section~\ref{sec:scratchpad}.

\begin{figure}[t]
\begin{minipage}[t]{0.50\linewidth}
\begin{minted}{python}
def tok_map(x, func):
    # tokenwise map
    return np.array([func(xi) for xi in x])

def seq_map(x , y, func):
    # tokenwise map over two sequences
    return np.array([func(xi, yi) for xi, yi in zip(x,y)])
    
def indices(x):
    return np.arange(len(x))

def full(x, const):
    return np.full_like(x, const)
    
def sel_width(A):
    # returns the number of selected keys
    # for each query in binary attention matrix A
    return np.dot(A, np.ones(len(A)))
                     
\end{minted}
\end{minipage}%
\hfill
\begin{minipage}[t]{0.45\linewidth}
\begin{minted}{python}
def select(k, q, pred):
    # constructs a causal binary attention matrix
    s = len(k)
    A = np.zeros((s, s), dtype=bool)
    for i in range(s):
        for j in range(i+1):
            A[i, j] = pred(k[j], q[i])
    return A

def aggr(A, v):
    # applies attention matrix A to value sequence v
    out = np.dot(A, v)
    norm = sel_width(A)
    return np.divide(out, norm, out=np.zeros_like(v),
        where=(norm != 0))
        
def kqv(k, q, v, pred):
    # convenience function which composes
    # select and aggr, like an attention layer
    return aggr(select(k, q, pred), v)
    
\end{minted}
\end{minipage}
\captionof{listing}{{\bf RASP.} Basic Numpy implementation of the RASP core functions.
See Listing~\ref{raspL-core} in the Appendix for the full RASP-L core, which is slightly more expressive.
}
\label{rasp:core}
\end{figure}

\begin{figure}
    \centering
    \includegraphics[width=\textwidth]{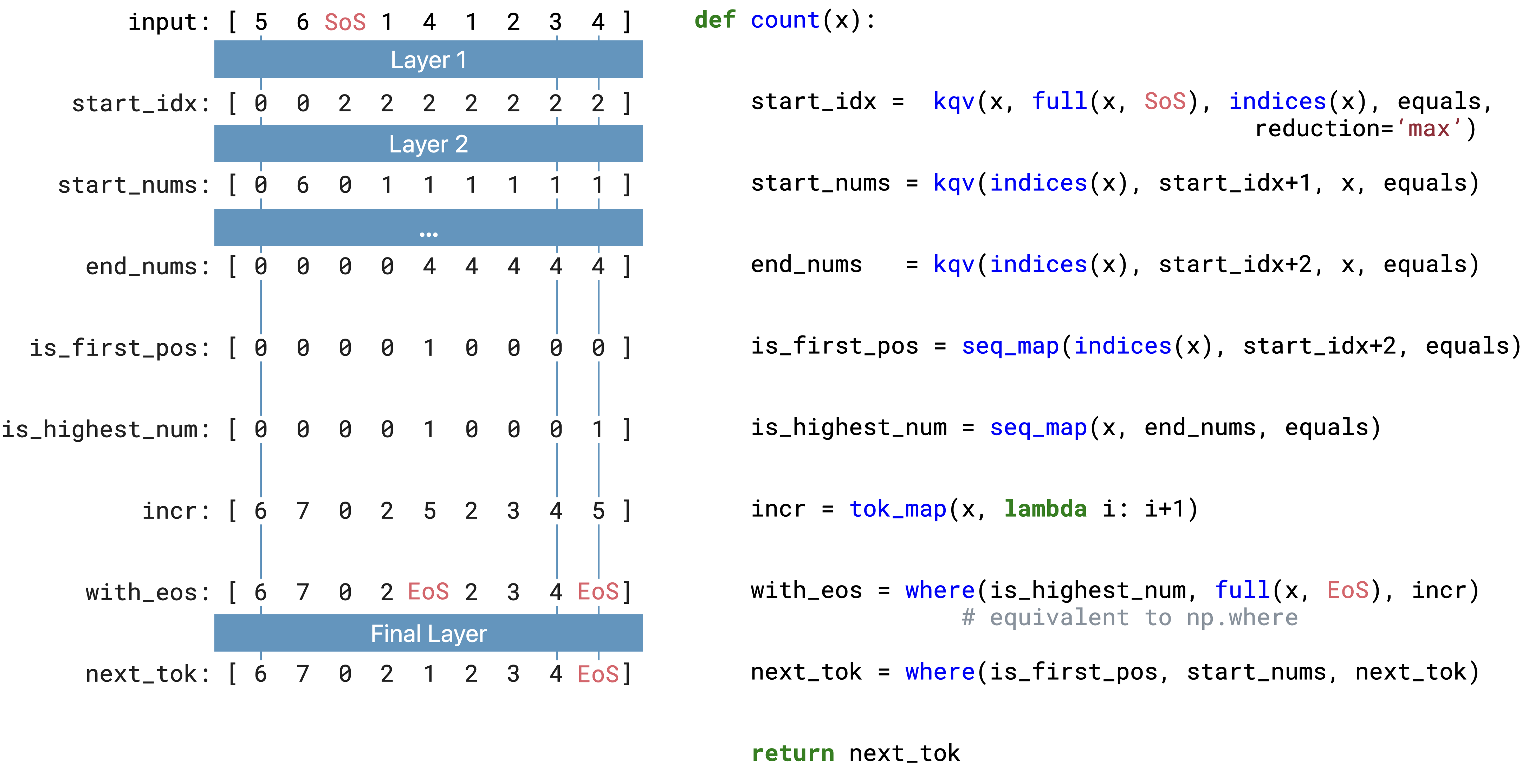}
    \caption{{\bf RASP-L program for Counting.}
    A RASP-L program that solves the counting task from the Introduction, along with an
    autoregressive Transformer that implements this program.
    Here the input $x$ is a context window that contains the prompt $[\textrm{SoS}, 1, 4]$,
    for counting from $1$ to $4$.
    For each input token, the program must output the correct next-token,
    just as an autoregressive Transformer would.
    We show all intermediate variables of the RASP-L program evaluated on $x$;
    note that these variables are all \emph{sequences},
    and since every RASP-L operation is causal,
    all intermediate variables are causal functions of the input as well.
    The \texttt{where} helper-function is implemented in the standard library of Listing~\ref{rasp:library}.
    We tokenise SoS as $-1$ and EoS as $-2$.
    }
    \label{fig:count-fig}
\end{figure}

\subsection{RASP-L: Learnable RASP}
\paragraph{Motivation.}
The original RASP language
defines a large set of programs which
can be represented by Transformers, by
hard-coding specific model weights.
However, there are two immediate problems with RASP
for our purposes: first, RASP is ``too expressive,'' and includes
programs which are in theory possible to represent, but
difficult to learn from scratch.
(For example, RASP allows arbitrarily-complex tokenwise operations
$\R \to \R$).
Second, the RASP representation is technically ``non-uniform\footnote{This is using the terminology from computational complexity; see for example \citet{arora2009computational}.}'', meaning
it requires a \emph{different} set of Transformer weights for each input length---
primarily because of how it handles indexing.
We thus introduce RASP-L, which is essentially a simplification of RASP
that is intended to be easy-to-learn and uniform.
In the experimental section, we will argue that RASP-L programs
can in fact be learned by Transformers, in a length-generalizing way.

To address these concerns,
we define a variant of RASP which we call RASP-L,
which is a restricted subset of RASP along with one additional feature.
The technical details of RASP-L are in Appendix~\ref{sec:rasp_spec}
but the primary restrictions are:
all variables are bounded integers (\bint) to avoid arbitrary-precision 
and numerical stability issues,
and token indices are treated specially.
Roughly, token indices in RASP-L can only be operated on in simple ways
(order comparison, predecessor, successor)--- arbitrary arithmetic involving
indices is not allowed. We will elaborate on the reasons for this below.
Finally, RASP-L has one added feature:
it allows ``min'' and ``max'' aggregations in the \texttt{aggr} function,
in addition to the standard ``mean'' aggregation of RASP.
These aggregations can be represented by an attention layer as well,
similar to mean-aggregation; we give an explicit construction in Appendix~\ref{sec:rasp-aggr}.

\paragraph{Example and Standard Library.}
In Figure~\ref{fig:count-fig} we walk through a detailed example of a \mbox{RASP-L} program
that solves the counting task from the Introduction.
To help write such programs,
in Appendix~\ref{app:rasp_programs}.\ref{rasp:library} we provide a small library of
useful helper-functions built on the RASP-L core.
This library includes, for example, an \texttt{induct} function
mirroring the ``induction heads'' identified by \citet{olsson2022context}.
The library notably does \emph{not} include the C function \texttt{atoi},
which parses a sequence of tokens as a decimal integer---
because operating on decimal representations is nontrivial:
it is not clear if Transformers can implement this parsing 
with a single algorithm that works for numbers of all lengths.

\paragraph{Index Operations.}
Our restrictions on token indices are
important because they rule out certain operations that seem natural in
common programming languages like Python,
but that would be challenging for a Transformer model to learn or represent.
For example, standard RASP allows arbitrary arithmetic operations on indices, such
as division-by-two: $i \mapsto \lfloor i/2 \rfloor$.
While such index operations can in theory be represented by Transformers,
this representation is not ``uniform'', meaning that input sequences of different lengths
may need entirely different positional embeddings.
Even in theory, it is non-trivial to construct
positional embeddings which encode all indices in $\N$,
while allowing arithmetic [ring] operations on indices
to be performed in an ``easy'' and numerically-stable way.
Thus we may intuitively expect that positional embeddings which were learned
to support index operations over say length 20 inputs will not remain valid
for length 50 inputs, because there is no ``natural'' representation
which works for both these lengths\footnote{We do not claim this is a fundamental limitation
of transformers, but it is a prominent feature of the standard experimental setting we consider.}.
To reflect this in RASP-L,
we only allow the following operations on indices:
order comparisons with other indices,
and computing successor/predecessor of an index.
This is formalized by a type system in Appendix~\ref{sec:rasp_spec}.

\section{Experiments: Length Generalization }
\label{sec:lengenexp}

In this section, we experimentally evaluate $4$ tasks that have simple RASP-L programs and $3$ tasks that do not.
We show that the tasks with simple RASP-L programs length-generalize well,
while the remaining tasks length-generalize poorly.
The four easy tasks we consider are: count, mode, copy with unique tokens, and sort. We provide the RASP-L program for these tasks in Appendix~\ref{app:rasp_programs}.
The three hard tasks we consider are: copy with repeat tokens, addition, and parity.

\subsection{Experimental Setup}
\label{sec:setup}
Our experimental setup is designed to mirror standard training procedure for LLMs as closely as possible.
Notably, this deviates from the typical setup used for synthetic tasks
in the following way:
at train time, we ``pack the context'',
filling the Transformer's context window with multiple independent samples
of the task, and we randomly shift the Transformer along its context window.
This procedure of packing and shifting the context mirrors standard practice in LLM
training \citep{nanoGPT,brown2020language},
but is typically not done in prior works using synthetic tasks.
It is an important detail: packing and shifting the context allows all positional embeddings to be trained,
and encourages the transformer to treat all positions symmetrically.
At test time, we evaluate examples without packing and shifting.
We measure the exact match (EM) on all outputs, which is $1$ if the entire output sequence is correct, and $0$ otherwise. 

All other experimental details are routine: we train decoder-only Transformers
with learned positional embeddings, trained from scratch, with the standard autoregressive loss on all tokens including the prompt.
We train all of our models to convergence on the train distribution where possible,
and we sample independent train batches from this distribution
(instead of using a finite-size train set).
For all tasks, the length of training examples is sampled uniformly from length $1$ up to the max training length.
The detailed experimental setup and hyperparameters are provided in Appendix~\ref{app:exp-details}.

\subsection{Successful Length Generalization}

\begin{figure}[t]
\captionsetup[subfigure]{justification=centering}
	\centering
	\subfloat[Length generalization on mode]{\includegraphics[width=.33
	\linewidth]{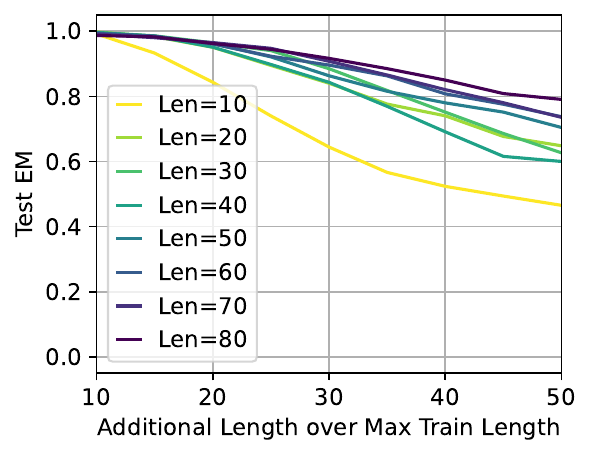}\label{fig:mode}}
	\subfloat[Copy task]{\includegraphics[width=.33\linewidth]{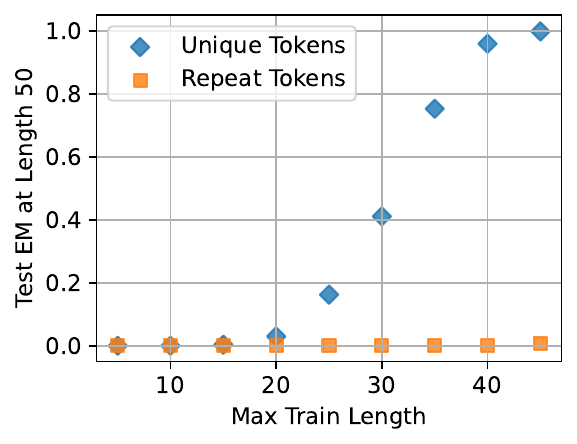}\label{fig:copy}}
 	\subfloat[Mode task (w/ scratchpad)]{\includegraphics[width=.33\linewidth]{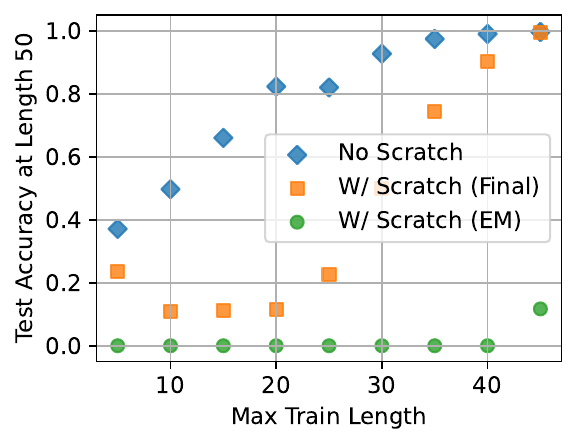}\label{fig:mode-scratch}}
	\caption{ 
     {\bf \protect\subref{fig:mode}} Length generalization
    performance for the mode task. All models generalize perfectly to $10$ or more additional tokens at test time. {\bf \protect\subref{fig:copy}} Performance on copy tasks for
    test sequences of length $50$, and varying train lengths.
    Copying unique tokens makes length generalization much easier,
    since it can be solved by an induction head.
    {\bf \protect\subref{fig:mode-scratch}} Performance on mode task
    for test sequences of length $50$, and varying training lengths.
    The scratchpad hurts in this setting (both final answer accuracy and exact match), since it is not computable by a short
    RASP-L program.
 }
	\label{fig:task-lengen}
\end{figure}

\paragraph{Count.} We described the count task in the Introduction and showed results in Figure~\ref{fig:count-lengen-scale}. The task outputs the sequence between a start and end token in the prompt, for example: \texttt{%
\setlength{\tabcolsep}{2pt}
\small
\begin{NiceTabular}{*{7}{c}}[hvlines]
2 & 5 & > & 2 & 3 & 4 & 5
\end{NiceTabular}}. Length here refers to the number of output tokens.
This task can be solved by the RASP-L program in Figure~\ref{fig:count-fig}.
We find that models trained on count can generalize near perfectly to double the training lengths.
It is crucial that our training distribution contain samples ranging from length $1$ to maximum training length,
which adds diversity as we scale up training length.
These factors are necessary in preventing shortcut programs from being learned. For example,
no generalization is observed if we train on sequences of all the same length. Moreover, as we scale up training length, we observe a corresponding increase in the number of tokens beyond the training length that the model can successfully generalize to.

\paragraph{Mode.} The mode task identifies the most frequent element in a sequence. \citet{merrill2022saturated} studied a binary-only version of this task and demonstrated strong length generalization.
We constrain the sequences such that the answer is unique. An example is:%
\texttt{%
\setlength{\tabcolsep}{2pt}
\small
\begin{NiceTabular}{*{10}{c}}[hvlines]
a & b & b & c & b & a & c & b & > & b
\end{NiceTabular}}. Figure~\ref{fig:mode} shows the results on mode,
when training and testing on random sequences from an alphabet of 52 symbols.
We find that models trained on mode generalize strongly to sequences far longer than the training lengths. Similar to count, we also observe here that increasing data diversity in the form of maximum training length leads to better length generalization. Interestingly, the improvement due to training set complexity in this task is much more subtle: even models trained on sequence length up to $10$ can achieve a median test accuracy of $50\%$ of sequences of length $60$.

\paragraph{Copy with unique tokens.} The copy task repeats the prompt sequence in the output.
We constrain the sequences to have all unique tokens in the prompt. An example is:%
\texttt{%
\setlength{\tabcolsep}{2pt}
\small
\begin{NiceTabular}{*{9}{c}}[hvlines]
a & c & d & b & > & a & c & d & b
\end{NiceTabular}}. Figure~\ref{fig:copy} shows the results on copy with unique tokens.
For models trained on sequence length up to $40$, we find that they can generalize perfectly to length of $50$.
Intuitively, this task is easy because we can leverage what is called an ``induction head'' \citep{olsson2022context}.
Induction heads work by identifying a previous instance of the current token,
finding the token that came after it,
and predicting the same completion to the current token.
\citet{olsson2022context} found that induction heads are reliably learned even by simple Transformers, and conjectured them to be a component of what enables in-context learning in LLMs.
Induction heads are simple to implement in RASP-L,
as the \verb|induct| function in Listing~\ref{rasp:library}. Thus, the next token
can be generated by simply using an induction head on the current token, since all tokens are unique.
This is exactly what the RASP-L program does, in Listing~\ref{rasp:copy_unique}.

\paragraph{Sort.} The sort task takes in a sequence and outputs the same sequence sorted in ascending order. For example: \texttt{%
\setlength{\tabcolsep}{2pt}
\small
\begin{NiceTabular}{*{9}{c}}[hvlines]
4 & 12 & 3 & 7 & > & 3 & 4 & 7 & 12
\end{NiceTabular}}. This task has been studied by \citet{li2023representations, awasthi2023improving}, and showed signs of strong length generalization.
Indeed, there is a one-line RASP-L program for this task (Listing~\ref{rasp:sort}).
Figure~\ref{fig:sort} shows the experimental results for sort. We observe strong length generalization on length $50$ for models trained with sequences of length $35$ or more.

\subsection{Unsuccessful Length Generalization}
Next, we study three tasks that do not admit simple RASP-L solutions: addition, parity,
and copy with repeating tokens.
We discuss reasons why Transformer models struggle to generalize on these tasks by highlighting the operations that these algorithms require, but that are unnatural for a Transformer to represent.
\paragraph{Addition \& Parity.}

Addition and parity have both been studied extensively as difficult tasks for Transformers. Models trained from scratch show little to no length generalization on addition \citep{nye2021show, lee2023teaching} and parity \citep{bhattamishra2020ability, chiang2022overcoming, ruoss-etal-2023-randomized, delétang2023neural}, and even pretrained LLMs
cannot solve these tasks robustly \citep{brown2020language, chowdhery2022palm, anil2022exploring} without careful prompting \citep{zhou2022teaching}. 

Indeed, addition is also difficult to write in RASP-L.
To see why, consider the standard addition program shown in Appendix~\ref{app:rasp_programs}.\ref{rasp:illegal_add}. 
This algorithm requires the carry value to be propagated in reverse order from least- to most-significant digit,
but this is difficult to simulate due to causal masking.
Moreover, the most prohibitive aspect is the index-related operations.
The standard addition algorithm requires index-arithmetic
(e.g. finding the middle of the prompt sequence) and precise indexing operations
(e.g. look up the corresponding summand digits for the current output digit).
Such operations are forbidden in RASP-L,
as they require index-arithmetic which are difficult to represent in a global, length-generalizing way.

Similarly, parity without any scratchpad requires operations that are forbidden under RASP-L.
The natural algorithm for parity on $n$ is to run a finite-state-machine for $n$ steps,
keeping the running-parity of prior bits as state.
However, arbitrary finite-state-machines cannot be
naturally simulated by one forward-pass of a transformer,
since one forward-pass involves only 
a constant number of ``parallel'' operations, not $n$ sequential ones.
Intuitively, solving parity ``in parallel'' requires taking
the sum of the entire sequence, then determining the parity of the sum.
This cannot naturally be computed in a numerically stable way for arbitrarily large sums.
Moreover, we cannot expect to learn a `sum' operation which
generalizes to numbers larger than the training sequences. 
Many works have shown that a Transformer cannot even fit the training set of parity sequences over some minimal length \citep{hahn2020theoretical, delétang2023neural}.

Under our experimental settings, we find that \emph{no} length generalization is observed
for both addition and pairity tasks:
test performance does not exceed random chance when
evaluated on examples $5$ or more tokens longer than the training set.

\paragraph{Copy with repeating tokens.}
For this task, we constrain the sequences to consist only of $2$ possible tokens.
An example is:%
\texttt{%
\setlength{\tabcolsep}{2pt}
\small
\begin{NiceTabular}{*{9}{c}}[hvlines]
a & a & b & a & > & a & a & b & a
\end{NiceTabular}}. Since the tokens are no longer unique, the induction-head is no longer helpful.
Instead, the model must perform precise index-arithmetic,
which are prohibited under RASP-L.
We show in Figure~\ref{fig:copy} that models fail to generalize to longer lengths on this task.

%% file: scratchpads.tex
\section{Application: Improving Length Generalization}
\label{sec:scratchpad}

In this section, we demonstrate how our RASP conjecture
can go beyond post-hoc explanations, by constructing interventions that predictably change length generalization performance. We study how reformatting tasks to allow shorter RASP-L programs can improve generalization performance, and how increasing diversity in the training data allows the model to perform well on tasks that require more complex RASP-L programs.

\subsection{Deep Dive on Addition}

\newcommand{\ta}{\textcolor{blue}{a}}
\newcommand{\tb}{\textcolor{purple}{b}}

\paragraph{Reducing the RASP-L complexity for addition.} In the previous section, we noted two aspects of a naive addition algorithm that pose problems for RASP-L: index-arithmetic (to query the summand digits for the current output digit), and non-causality (for the carry).
To address the difficulty with indexing operations, we can leverage induction heads to simplify the addition algorithm for a Transformer
by adding ``index hints'' to the prompt and answer:
For example, \texttt{%
\setlength{\tabcolsep}{1pt}
\small
\begin{NiceTabular}{*{8}{c}}[hvlines]
5 & 4 & + & 3 & 7 & > & 9 & 1
\end{NiceTabular}} becomes \texttt{%
\setlength{\tabcolsep}{1pt}
\small
\begin{NiceTabular}{*{14}{c}}[hvlines]
\ta & 5 & \tb & 4 & + & \ta & 3 & \tb & 7 & > & \ta & 9 & \tb & 1
\end{NiceTabular}}.
This enables us to get the corresponding digits
for each sum step by calling \verb|induct| on its index hint
(\texttt{\ta} or \texttt{\tb}),
thus sidestepping the need to precisely access and manipulate positional indices.
During training, we sample the index hints as a random slice from a longer
contiguous block of tokens, to encourage learning all hints and their linear ordering.
This is similar to our training strategy for the count task.
Adding index hints thus allows us to avoid index-arithmetic, which is the most prohibitive aspect of representing the addition algorithm in RASP-L.

\begin{figure}
    \centering
    \includegraphics[width=.6\linewidth]{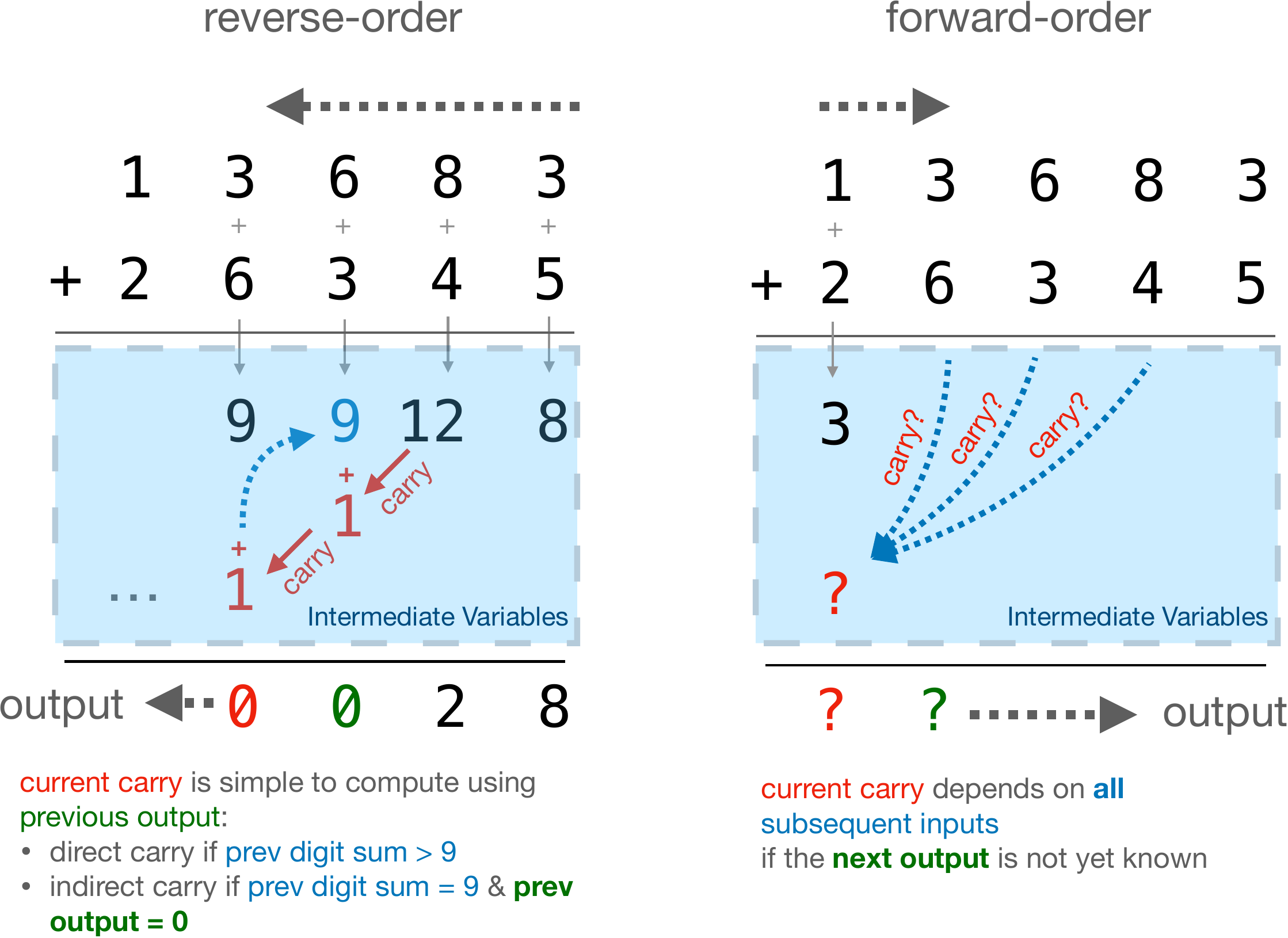}
    \caption{Intuition for why reverse order addition is simpler to implement in RASP than forward order. In reverse order, the carry calculation is simple with the help of the most recent output digit, which acts like a built-in scratchpad. In forward order, the carry requires a more complex calculation involving all remaining input digits.
    This intuition is reflected in the RASP-L program for forward addition,
    which is much longer than that of reverse addition (see Listings~\ref{rasp:add_reverse_index_hints} and \ref{rasp:fwd-block}).
    }
    \label{fig:reverse_add_diagram}
\end{figure}

To address non-causality of the carry operation,
we can format the output in reverse-order (from least- to most-significant digit).
For example, \texttt{%
\setlength{\tabcolsep}{1pt}
\small
\begin{NiceTabular}{*{14}{c}}[hvlines]
\ta & 5 & \tb & 4 & + & \ta & 3 & \tb & 7 & > & \ta & 9 & \tb & 1
\end{NiceTabular}} becomes \texttt{%
\setlength{\tabcolsep}{1pt}
\small
\begin{NiceTabular}{*{14}{c}}[hvlines]
\ta & 5 & \tb & 4 & + & \ta & 3 & \tb & 7 & > & \tb & 1 & \ta & 9
\end{NiceTabular}}.
This enables simple and causal propagation of the carry where each step can reference
the previous output to determine the current carry, similar to the standard addition algorithm.
A RASP-L program for reverse-order addition, with index-hints,
is provided in Listing~\ref{rasp:add_reverse_index_hints}.
Although reversing the answer digits greatly simplifies the carrying procedure,
it is still possible to implement an algorithm for addition in standard order.
This algorithm is nontrivial,
because we essentially need to propagate 
a carry through all $n$ input digits
in order to determine the first digit of output, the most significant digit
(see Figure~\ref{fig:reverse_add_diagram}). Although arbitrary sequential operations cannot be implemented in a Transformer, we can parallelize these operations due to their simplicity.
We show how to construct a RASP-L program
for standard-order addition in Listing~\ref{rasp:fwd-block}.
Comparing Listing~\ref{rasp:add_reverse_index_hints} and Listing~\ref{rasp:fwd-block}
reveals how much more complicated the forward algorithm is--- it results in a much longer RASP-L program.
The observation that reverse-order addition is easier for autoregressive Transformers
was made in \citet{lee2023teaching}, and is usually
explained by claiming the reverse-order algorithm is ``simpler'' than the forward-order one.
Here we make this explanation more concrete, by 
using a notion of ``simpler'' that is specific to the Transformer's computational model
(that is, RASP-L program length).

\begin{figure}[t]
\captionsetup[subfigure]{justification=centering}
	\centering
	\subfloat[Forward-order addition with index hints]{\includegraphics[width=.33
	\linewidth]{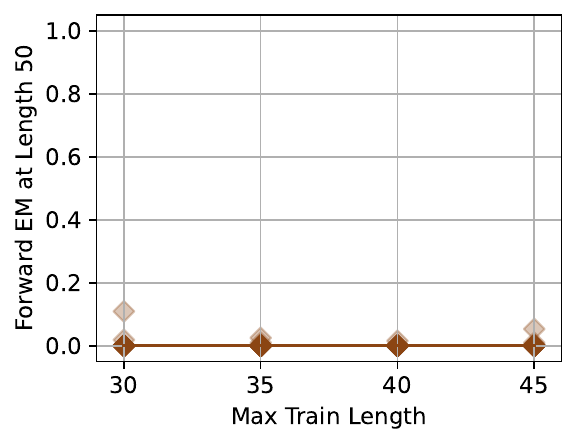}\label{fig:addforward}}
	\subfloat[Reverse-order addition with index hints]{\includegraphics[width=.33\linewidth]{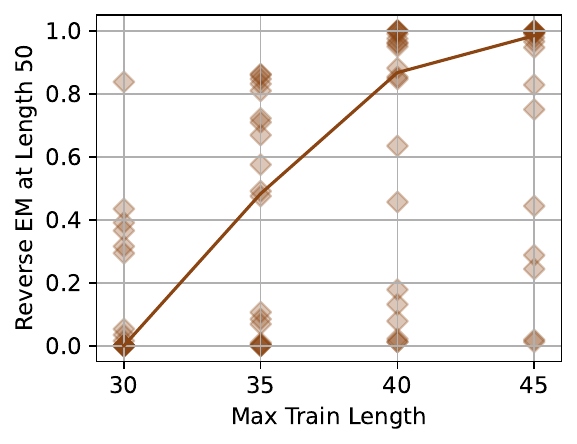}\label{fig:addrev}}
 	\subfloat[Parity with scratchpad]{\includegraphics[width=.33\linewidth]{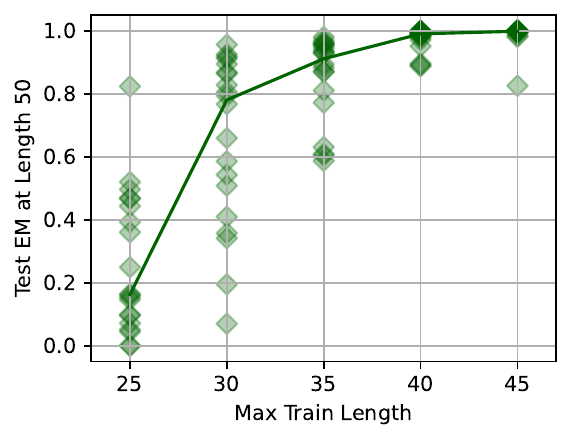}\label{fig:parity}}
	\caption{{\bf Length generalization on addition and parity.}
 Plot shows $20$ individual trials per length, as well as their median (solid line).
 {\bf \protect\subref{fig:addforward}} Shows generalization performance for forward addition with index hints on hard carry examples of length $50$. No length generalization is observed. {\bf \protect\subref{fig:addrev}} Shows
 the generalization of reverse addition with index hints on hard carry examples of length $50$.
 Most runs start to length generalize perfectly on $50$ digit addition once the training length is greater than $40$.
    {\bf \protect\subref{fig:parity}} Shows generalization performance for parity with scratchpad, on length $50$ inputs.
    Most runs start to generalize perfectly on $50$ digit addition once the training length is greater than $35$. 
 }
	\label{fig:add-parity-scratch}

\end{figure}

\paragraph{Index hints enables generalization on addition.} We evaluate addition in two settings: ``easy'' carry and ``hard'' carry.
In easy carry, the two numbers are sampled randomly and independently--- this is what is typically done in the literature.
However, uniformly random summands will only produce addition
instances with short carry-chains (in expectation)---
and for such instances, each output digit only depends on a small number of input digits.
We thus also test ``hard'' carry instances, where we constrain the
examples to have the longest possible carry chain for the given length.
For example, a hard carry instance of length $3$ is $381 + 619 = 1000$,
which requires the model to
compute the carry over a chain of $3$ digit positions.
The performance on ``easy'' carry is shown in Figure~\ref{fig:addeasy},
and the performance on ``hard'' carry in Figure~\ref{fig:add-parity-scratch}.
We find that index hints allow both forward
and reverse addition to length generalize on ``easy'' carry.
However, on ``hard'' carry questions that 
involve carry-chains longer than seen at training,
reverse addition maintains strong length generalization
while forward addition exhibits no generalization. 
Moreover, we observe in Figure~\ref{fig:addtrainspeed} that reverse addition optimizes more quickly than forward addition during training.
These observations are all consistent with our claim that
length generalization is ``easier''
on tasks which admit simple RASP-L programs.

\paragraph{Diversity enables generalization on forward addition.}
As we saw previously, one lever to improve generalization is to convert the task into one that has a simpler RASP-L program. Another lever suggested by the RASP conjecture is to increase training data diversity, such that shortcut programs can no longer fit the training set.
Since forward addition does admit a RASP-L program, albeit a more complex one,
we would expect it is possible to learn if we ``try harder,'' e.g. use a more careful
and diverse train distribution.
We explore this by training with balanced carry sampling--- instead of sampling the two numbers independently, we first sample the length of the carry chain uniformly between 0 and question length, then sample a random question that contains the given carry chain length. This ensures that the model sees a significant percentage of questions containing long carry chains, thus increasing the diversity and difficulty of the training data.
The results of the balanced carry training approach for both forward and reverse addition are shown in Figure~\ref{fig:addbalanced}. We see that this more careful training unlocks the model's ability to length generalize on forward addition, even under the hard carry evaluation.
To our knowledge, these results demonstrate the first instance of strong length generalization
on decimal addition for Transformer models trained from scratch.

\subsection{Why do Scratchpads Help?}

The RASP conjecture provides a natural way to understand
why scratchpads \citep{nye2021show,wei2022chain} can be helpful:
scratchpads can simplify the next-token prediction task,
making it amenable to a short RASP-L program.
One especially common type of simplification is when a scratchpad is used to
``unroll'' a loop, turning a next-token problem that requires $n$ sequential steps
into $n$ next-token problems that are each only one step.
The intuition here is that Transformers 
can only update their \emph{internal} state in very restricted ways
---given by the structure of attention--- but they can update their \emph{external} state (i.e. context) in much more powerful ways.
This helps explain why parity does not have a RASP-L program, but addition with index hints does.
Both tasks require some form of sequential iteration,
but in the case of addition, the iteration's state 
is \emph{external}: it can be decoded from the input context itself.

In the following examples, we construct ``good'' scratchpads 
which convert the original problem into one that can be solved by a simple RASP-L program.
Conversely, we construct ``bad'' scratchpads which seem natural to humans but require a more complex RASP-L program than the original task. We show that these interventions lead to predictable changes in length generalization, per the RASP conjecture.

\paragraph{Scratchpad enables generalization on parity.}

The natural algorithm for parity is to iterate through all input tokens,
updating some internal state along the way. 
Unfortunately, this algorithm cannot be directly implemented in RASP-L,
because RASP-L does not allow loops--- fundamentally
because one forward-pass of a Transformer cannot directly simulate
$n$ sequential iterations of the loop.
However, if the internal state is written on the scratchpad each iteration,
then the Transformer only needs to simulate \emph{one} iteration in one forward-pass,
which is now possible.

\begin{figure}[t]
\captionsetup[subfigure]{justification=centering}
	\centering
	\subfloat[Training speed comparison for addition with index hints]{\includegraphics[width=.33
	\linewidth]{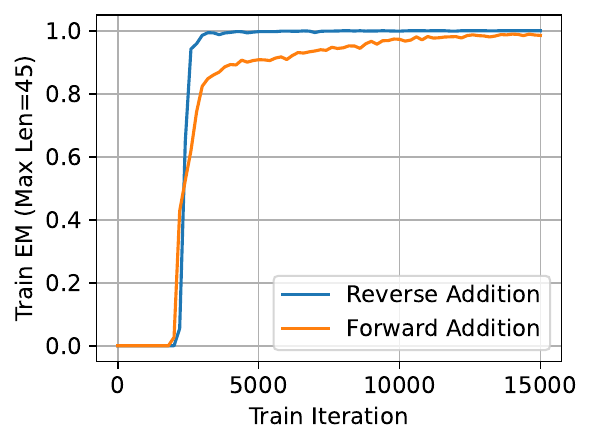}\label{fig:addtrainspeed}}
	\subfloat[Training speed comparison for parity with different scratchpads]{\includegraphics[width=.33\linewidth]{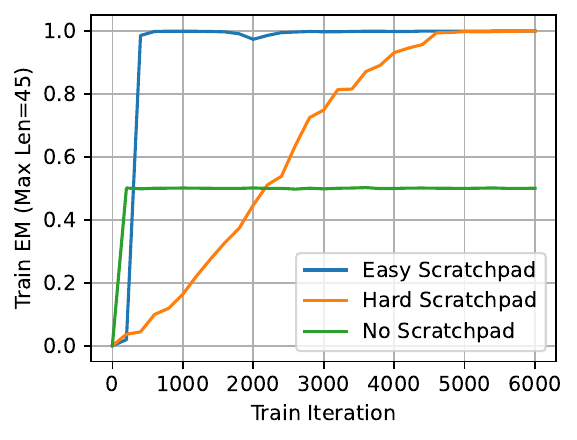}\label{fig:partrainspeed}}
 	\subfloat[Training speed comparison for mode with scratchpad]{\includegraphics[width=.33\linewidth]{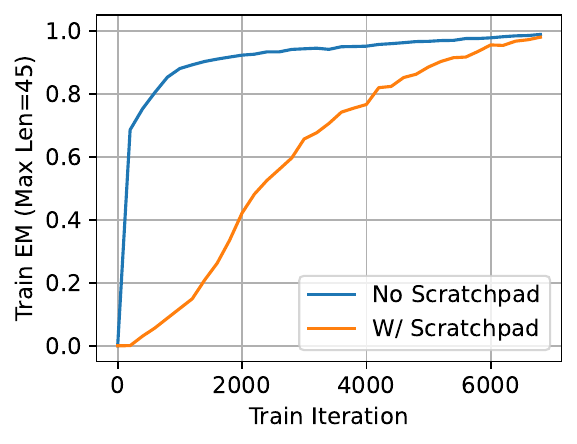}\label{fig:modetrainspeed}}
	\caption{We compare the training speed as measured by exact match on a maximum training length of $45$. {\bf \protect\subref{fig:addtrainspeed}} compares the convergence speed of models trained on forward addition vs reverse addition with index hints. {\bf \protect\subref{fig:partrainspeed}} compares the convergence speed of models trained on different scratchpads on parity.
 }
	\label{fig:trainspeedlengen}

\end{figure}

We leverage this intuition to design a scratchpad for parity. Similar to addition, we add index hints to the prompt to simplify the indexing operation.
In the scratchpad output, we locate index hints that precede each $1$ in the prompt,
and keep track of the running parity with symbols $+$ (even) and $-$ (odd).
The last output token corresponds to the final answer. For example:\texttt{%
\setlength{\tabcolsep}{1pt}
\small
\begin{NiceTabular}{*{16}{c}}[hvlines]
a & 0 & b & 0 & c & 1 & d & 1 & e & 0 & > & + & c & - & d & +
\end{NiceTabular}}. Figure~\ref{fig:parity} shows the exact match performance of the proposed parity scratchpad. We find that some of the runs trained with sequences up to $30$ in length can generalize perfectly on sequences of length $50$. When training length reaches $40$, all models achieve perfect length generalization on length $50$.
Lastly, in Figure~\ref{fig:partrainspeed}, we compare the training curves of parity using various scratchpads, and we show that training speed is also correlated with how difficult the task is under RASP-L. Details can be found in Appendix~\ref{app:trainspeed}.
To our knowledge,
these results demonstrate the first instance of strong length generalization on parity for Transformer models trained from scratch.

\paragraph{Scratchpad \textit{hurts} generalization on mode.}

Now we consider the mode task and look at how scratchpad might affect a task that a Transformer is naturally amenable to. A natural algorithm one might come up with is to calculate the counts of each unique token in the sequence, then output the token with the maximum count.
To encourage the model to learn this algorithm, we might utilize the following scratchpad,
where we output the frequency of each token in ascending order:%
\texttt{
\setlength{\tabcolsep}{1pt}
\small
\begin{NiceTabular}{*{16}{c}}[hvlines]
a & b & b & c & b & a & c & b & > & 2 & a & 2 & c & 4 & b & b
\end{NiceTabular}}. 
The last token in the scratchpad is then the correct answer.
However, although this scratchpad provides more supervision for what algorithm the model should learn,
it is a more difficult task when considered in terms of RASP-L.
Intuitively, instead of internally counting element-frequencies and outputting the maximum,
the model must now explicitly sort the frequencies, and convert
these into decimal representations to produce its scratchpad.

We show in Figure~\ref{fig:mode-scratch} that the scratchpad performs significantly worse than no scratchpad, both when measured on exact match and also on the accuracy of the final answer. This shows that not only is the scratchpad difficult to generalize on (low EM), but it also reduces the likelihood that the final token corresponding to the answer learns the correct solution. Moreover, we show in Figure~\ref{fig:modetrainspeed} that models trained on scratchpad converges much more slowly during training than no scratchpad.
In Appendix~\ref{app:modescratch}, we evaluate another variant of the scratchpad with the tokens presented in order of appearance, such that no sorting is required, and observe a similar effect.

%% file: gotu.tex
\section{Comparison to Min-Degree-Interpolators}
\label{sec:gotu}
An essential aspect of our work is our
Transformer-specific notion of
function complexity: the minimum RASP-L program length.
Here we show why this choice is important, by contrasting it with another
popular notion of complexity: minimum polynomial degree.
Concretely, \citet{abbe2023generalization} recently proposed a model of learning
in which Transformers learn the minimum-degree function which interpolates their train set.
We will give a simple example where
our RASP toy model correctly predicts a Transformer's
out-of-distribution generalization behavior, but the min-degree-interpolator model
does not.
We emphasize that these results are not inconsistent with \citet{abbe2023generalization}:
neither \citet{abbe2023generalization} nor our current work claim to apply in all settings.
Rather, this section illustrates how a Transformer-specific measure of complexity
can be more predictive than architecture-agnostic measures of complexity, in certain settings.

\subsection{The Setting: Boolean Conjunction}
We consider the ``Generalization-on-the-Unseen'' setting of \citet{abbe2023generalization},
for direct comparison.
Our target function is simply boolean conjunction.
Given $n=20$ input bits $x_i \in \{0, 1\}$, the ground truth function
is the boolean AND of all bits:
$f^*(x_1, x_2, \dots, x_{n}) = \bigwedge_{i \in [n]} x_i$.
That is, $f^*(x) = 1$ iff all bits $x_i=1$, and $f^*(x) = 0$ otherwise.
Now, the train distribution is supported on inputs $x$ where the last $k=5$ bits of $x$ are always $1$.
Specifically: with probability $\nicefrac{1}{2}$, $x$ is the all-ones vector, 
otherwise $x$ is all ones with a single $0$ in a random location among the first 15 bits.
That is,
\[
x \sim (1^n - B(\nicefrac{1}{2})e_i); \quad i \sim \mathrm{Unif}[1, 15].
\]
where $e_i \in \{0, 1\}^n$ is the $i$th standard basis vector.
Note the ground-truth label $y = f^*(x)$ for this distribution is balanced.
The unseen test distribution is
identical to the train distribution, except the `0' is only among the last $5$ bits.
That is,
\[
x^{\textrm{(Unseen)}} \sim (1^n - B(\nicefrac{1}{2})e_i); \quad i \sim \mathrm{Unif}[16, 20].
\]
We now ask: when a Transformer is trained on the above train distribution,
what does it predict on the unseen test distribution?

\paragraph{Experimental Result.}
We train a standard decoder-only Transformer autoregressively in the above setting,
with sequence distribution $[x, f^*(x)]$ for $x$ sampled
from the train distribution.
The trained Transformer reaches 100\% test accuracy on the unseen test set.
That is, the Transformer
correctly computes the boolean AND, even on bits which were irrelevant at train time. 
Experimental details can be found in Appendix~\ref{app:exp-details}.

\subsection{The Min-Degree Interpolator}
We now claim that the minimum-degree-interpolator of the train set
\emph{does not} behave like the Transformer in the above experiment.
To see this, observe that the minimum-degree-interpolator will not depend on the last $k=5$
bits of the input, since these bits are constant on the train set.
This can be formalized via the following simple lemma, which 
uses the same notion of ``degree profile'' (\textrm{DegP}) as \citet{abbe2023generalization}\footnote{
Briefly, the degree profile of a boolean function is the tuple of its Fourier weights at each level,
with the natural total ordering which refines the standard polynomial degree.
}.

\begin{lemma}
\label{lem:mindegree}
For all subsets $S \subseteq \{0, 1\}^n$
and all boolean functions $f: \{0, 1\}^n \to \R$,
the following holds.
Let $g^*: \{0, 1\}^n \to \R$ be the boolean function of minimum degree-profile
which agrees with $f$ on $S$.
That is,
\[
g^* = \argmin_{\substack{g: \{0, 1\}^n \to\R \\ \textrm{s.t. } g\vert_S = f\vert_S}} \mathrm{DegP}(g).
\]
Let $I \subseteq [n]$ be the subset of indices (if any) on which $S$ is constant.
That is, $\pi_I(S)$, the projection of $S$ to coordinates $I$, is a singleton.
Then, the minimum-degree-interpolator $g^*$ also does not depend on indices $I$. That is,
\[
x_i = y_i \forall i \not\in I \implies g^*(x) = g^*(y).
\]
\end{lemma}

This lemma follows from the fact that if an interpolator depends on
bits which are always constant on the train set $S$,
then its degree-profile could be reduced without changing its output on $S$,
and thus it cannot be a min-degree-interpolator.
For completeness, we state this formally as Lemma~\ref{lem:helper} in the Appendix.

\paragraph{RASP-Length.}
On the other hand, there is a one-line RASP-L program which computes the boolean AND of all input bits:\\
\hspace*{\fill} {\small \verb|def output(x): kqv(x, full(x, 0), full(x, 0), equals, default=1)|} \hspace*{\fill}\\
This program can be represented by a 1-layer Transformer, where the attention layer simply ``searches for a $0$''
in the prior context.
Thus, our RASP toy model of learning predicts the correct experimental result in this setting.
The intuition here is that it is actually ``easier'' for a Transformer to represent
the boolean conjection of \emph{all} input bits, and so it learns to do so.
Learning to ignore the last $5$ bits would have been possible,
but is an additional complication, and does not help in fitting the train set in any case.
Notably, this intuition (and indeed, the experimental result) does not hold for MLPs; 
it is specific to the Transformer architecture.

%% file: discussion.tex
\section{Discussion, Limitations, and Open Questions}

\paragraph{Limitations of RASP-L.}
The RASP and RASP-L programming languages were remarkably useful
in our work,
for reasoning about which algorithms
are easy to represent and learn.
However, they have clear limitations. 
For one, RASP and RASP-L are not \emph{complete}: not all functions which
can be efficiently represented by Transformers have efficient RASP implementations.
A notable class of Transformer algorithms which are ill-suited for RASP are
numerical algorithms, such as in-context gradient-descent 
and linear regression, which involve high-dimensional
matrix and vector floating-point operations \citep{akyurek2022learning,garg2022can,charton2021linear}.
Moreover, RASP supports only deterministic outputs and binary attention masks. 
In addition, some of the RASP-L restrictions 
may seem somewhat contrived (such as no floating-points and index restrictions)--- there may be a more natural way to 
capture the set of easily-representable algorithms for standard Transformers.
Nonetheless, we view RASP and RASP-L as important steps
towards reasoning about Transformer algorithms, and we hope
to see future work in this direction.

\paragraph{Limitations of Complexity Measures.}
We considered for simplicity a basic notion of function complexity,
which is the minimum RASP-L program length.
This intuitive notion captures many empirical behaviors, as we demonstrated.
However, depending on the operations, it can be possible to represent multiple lines of RASP-L
in a single layer of a Transformer, and so RASP program length does not perfectly correspond to Transformer-complexity.
There are likely more refined notions of complexity, such as the ``parallel depth'' of RASP-L programs, or lower-level measures like the minimum-weight-norm among all
weight-noise-robust Transformers which implement the function.
Many of these notions, including ours, have the drawback of being likely intractable to compute---
intuitivly, it may be difficult to find the minimum RASP-L program for a task
for similar reasons that Kolmogorov complexity is uncomputable \citep{kolmogorov1963tables}.
We leave investigating more refined notions of complexity to future works.

\paragraph{Limitations of Scope and Strength.}
We acknowledge that our Main Conjecture is not fully formal,
because there are aspects we do not fully understand.
For example, we cannot precisely predict the \emph{extent}
of length generalization for different tasks.
Moreover, since it is likely intractable to determine
the minimum RASP-L program that fits a given training set,
we cannot predict a priori what forms of ``data diversity'' are required to ensure strong length generalization, even if our conjecture holds true.
Nonetheless, we view our conjecture as a step forward
in understanding the implicit bias of Transformers,
as it has more predictive power than many prior theories.
Developing more formal and precise conjectures is an important question for future work.

\paragraph{On Formal Language Characterizations.}
One important question that remains open is whether
there exists a natural complexity-theoretic definition of
the class of tasks which are ``simple'' for autoregressive
Transformers to represent.
For example, it is well-known that RNNs tend to length generalize
on tasks equivalent to \emph{regular languages}, like Parity
(e.g. \citet{delétang2023neural}).
This is intuitively because
regular languages can be decided by a class of algorithms
(deterministic finite automata) which are ``simple'' to represent by RNNs,
and thus plausibly easy to learn.
We would like an analogous characterization of which tasks
admit algorithms with a simple and natural Transformer representation.
Recent works have characterized which functions are
\emph{possible} to represent by a Transformer\footnote{Note that
the exact set of which functions are representable
depends on certain definitional details of Transformers such as finite vs. infinite precision, bounded vs. unbounded weights, etc, which is why some of these references arrive at different conclusions.},
but this representation is not always ``simple'' enough to be learnable, 
and not always uniform
\citep{hahn2020theoretical,merrill2022saturated,chiang2022overcoming,perez2021attention,bhattamishra2020ability,ebrahimi2020can}.
Our presentation of RASP-L is meant to be one way of defining algorithms which are ``simple'' to represent---
those expressable as short RASP-L programs---
but this definition is not explicit (and likely not complete).

\paragraph{Relations to Mechanistic Interpretability.}
The line of work on mechanistic interpretability 
seeks to understand which algorithms (or ``circuits'') Transformers
internally learn to implement, by inspecting their internals \citep{olsson2022context,nanda2023progress, zhong2023clock, conmy2023automated, zhang2023transformers, hanna2023does}.
Our work is thematically related, but differs in scope and level of abstraction.
Our toy model is a mechanistic model and provides intuition for the RASP conjecture, but we focus the scope of this paper on the phenomenological level. We aim to establish the predictive power of the conjecture in this work, and thus focus on the external performance of the trained models in the out-of-distribution setting;
whether trained Transformers'
internals are at all similar to the internals of a compiled RASP program remains an open question (some such investigation was done in \citet{lindner2023tracr}).
Finally, our motivation can be seen as dual to mechanistic interpretability:
while interpretability typically takes models which are known to ``work'' experimentally, and then investigates how they work,
we attempt to predict whether models will work or not before training them on a given task.

%% file: related.tex
\section{Additional Related Works}
\label{app:relatedworks}

Our paper is related to the line of work that seeks to understand the capabilities and limitations of Transformer models when it comes to algorithmic reasoning \citep{kaiser2015neural, velivckovic2021neural}. Specifically, we focus on simple tasks like arithmetic and study length generalization on the standard Transformer architecture. Related to this, \citet{lee2023teaching} study how well Transformers trained from scratch can learn simple arithmetic tasks, and finds that no length generalization is observed. \citet{nogueira2021investigating} find that partial length generalization on addition is observed only when models reach 3B parameters and when the addition questions are presented in reverse order.
\citet{jelassi2023length} study models trained on addition and find strong generalization performance when using a few examples of longer sequences. However, they required non-standard architectures (non-causal Transformers)
and training procedures (artificial padding) to observe this, and still find that the model does not generalize to unseen length that is in between the minimum and maximum lengths seen during training. 

Moreover, our work contributes to the study of what makes for effective scratchpads. Other papers have also found that using positional tokens in the prompt can help with length generalization \citep{nogueira2021investigating, li2023representations}. However, these works do not provide a framework for understanding why these tricks are helpful. A number of papers also study how chain-of-thought style prompting helps with reasoning performance \citep{wei2022chain, zhou2022teaching, zhou2022least, creswell2022faithful, madaan2022text}, but these focus on in-context learning and do not study the effect of training models on these formats.

Other papers also aim to understand the limits of what Transformers can learn and represent. \citet{bhattamishra2020ability} and \citet{delétang2023neural} study the ability of Transformers to represent and generalize on families of formal languages.
\citet{zhang2023transformers} evaluate the ability of Transformer models to emulate the behavior of structurally recursive functions from input-output examples. \citet{liu2023transformers2} study how shallow Transformers can simulate recurrent dynamics representable by finite-state automata. Both works identify shortcut solutions that become brittle on out-of-distribution samples. \citet{bhattamishra2023simplicity} suggest that Transformer models have an inductive bias towards learning functions with low sensitivity, such as sparse boolean functions, but focus on the in-distribution setting. \citet{abbe2023generalization} also propose a simplicity bias in Transformers, but use “minimum-degree” as their notion of function simplicity. However, they only consider functions with fixed input dimension rather than programs on arbitrary input lengths.

Lastly, there are other approaches to improving length generalization in Transformers. These include studying how various training hyperparameters and design choices influence compositional generalization \citep{furrer2021compositional, ontañón2022making}, and designing better positional embeddings \citep{press2022train, ontañón2022making, kazemnejad2023impact, ruoss-etal-2023-randomized}.

\paragraph{Relations to ``Faith and Fate.''}
At first glance, our results may seem in tension with the recent work
of \citet{FF}, which gave evidence that Transformers 
predict by ``subgraph matching'' parts of the
computational graph from their train sets,
without systematically learning the correct algorithm.
In contrast, our work presents multiple settings where 
Transformers do actually seem to learn the correct algorithm.
The resolution here is that whether Transformers compositionally generalize
or not depends crucially on the structure of the task,
as we have discussed.
Specifically, the notion of ``computational graph'' from \citet{FF}
implicitly depends on the computational model which executes the graph.
A key difference of our work is we distinguish the computational model of 
the Transformer architecture from that of the standard von Neumann architecture.
As we have seen, certain programs have complex computational graphs
when written in C, to run on a CPU (such as sorting algorithms),
but have much simpler computational graphs when written in RASP, to run on a Transformer --- and vice versa.
On a more minor note, we do not require models to
\emph{memorize} steps in the computation subgraph
from their training data, as in \citet{FF}.
Rather, we allow them to \emph{learn} steps in the appropriate subgraph,
which can be much more sample-efficient than memorization for certain
operations.
Finally, we agree with \citet{FF} that there is a fundamental ``noise floor''
to error-free length-generalization of autoregressive models. 
However,  it is theoretically possible to reduce the error-propogation
by ``scaling-up'' various axes (the data, model, train time, etc),
and indeed we observe quite strong length-generalization even in our 
relatively small experiments.

\paragraph{Transformers are RNNs.}
Mostly as a curiosity, we observe that there are likely even stronger computational
restrictions on RASP programs than on generic Transformers, for the following reason.
RASP programs can be compiled into standard Transformers,
but they can also be compiled into ``linear-attention Transformers'' --- Transformers
without the attention softmax ---
by essentially the same construction.
Linear-attention Transformers can evaluate their next-token function in 
$\mathcal{O}(n)$ time in their input length, as observed by \citet{katharopoulos2020transformers}.
Thus, RASP programs can only solve tasks where the next-token function
can be computed in linear-time (as opposed to the $\mathcal{O}(n^2)$ upper bound on generic Transformers).
This limitation technically may not apply to RASP-L, since we allow for ``max/min'' aggregation
types that require the softmax to implement.
However, this restriction applies to any RASP-L program
which does not use these types of aggregation.

\paragraph{Insights on Addition.}
Our framework gives one way to understand some of the intriguing experimental
observations of \citet{lee2023teaching}.
For example, consider the results of Section 5.2 in \citet{lee2023teaching}.
Briefly, they train a Transformer on 3-digit decimal addition problems,
but they withhold the digit ``5'' from appearing as the first digit
of numbers in the train set (the digit ``5'' appears in other positions).
They then test on 3-digit addition problems with a ``5'' as the first digit, and 
find remarkably nontrivial generalization.
This is apriori fairly surprising, 
as \citet{lee2023teaching} observes,
and cannot be explained by certain simple models of learning such as low-rank matrix completion.
From our perspective, one potential way to understand this is that the Transformer
must learn the same program to predict the first output digit
as for the second and third output digits, and the ``simplest'' such program
to represent for a Transformer is one which treats all digits symmetrically.

%% file: appendix.tex
\section{Additional Experimental Details}
\label{app:exp-details}

For all experiments, we tokenize every character individually, including digits of a number (unless otherwise specified). We train in the online setting, where each batch is sampled iid from the train distribution instead of from a finite train set --- this avoids overfitting issues, and is closer to the training of modern LLMs. Unless otherwise specified, we evaluate test performance on $5 \times$ the batch size number of samples. Unless otherwise specified, we run each experiment on $20$ random seeds and report the median of the runs. We select hyperparameters for each task based on what is required to fit the training set. Hyperparameter details can be found in Table~\ref{tab:exp_setting}. 

\paragraph{Count.}
For the count task, we train with an alphabet size of $155$ and evaluate on test sequences up to $150$ in length. 
For this particular task, we tokenize each number as its own token, without considering digits--- that is, there
is one token for every integer $i \in [0, 155)$.
Given the nature of the task, we enumerate all possible sequences of each test length at evaluation time.
At train time, the length of each example is sampled uniformly between $1$ and the maximum training length.

\paragraph{Mode.} For the mode task, we train on an alphabet of $52$ tokens. Each example consist of $5$ unique tokens sampled randomly from the alphabet. The length of each example is sampled uniformly between $1$ and the maximum training length, and the sequence is randomly sampled from the $5$ selected tokens. If there is a tie for the most frequent element, we randomly select a token from one set and changes it to a token from the other set, thus ensuring that there is one unique answer.

\paragraph{Copy.} For the copy task with unique tokens, we train on an alphabet size of $100$. The length of each example is sampled uniformly between $1$ and the maximum training length, and the sequence is randomly sampled from the alphabet without replacement. For the copy task with repeat tokens, we use the same sampling procedure, but now on an alphabet size of $2$. 

\paragraph{Sort.} For the sort task, we train on an alphabet of $100$ tokens. The length of each example is sampled uniformly between $1$ and the maximum training length, and the sequence is randomly sampled from the alphabet without replacement. 

\paragraph{Addition.} For the addition task, all length of numbers are sampled uniformly from $1$ up to the maximum training length, and numbers are then sampled conditioned on this length. In ``standard carry'' setting, we sample the length of each of the two numbers independently. In ``balanced carry'' setting, the length of the carry chain is sampled uniformly from $0$ up to the length of the first number, and the position of the carry chain is randomly selected as a segment of the first number; then the second number is sampled based on the required carry chain. We then pad the two numbers with $0$ in the front such that they have the same length. We pad the numbers with an extra $0$ to allow for the potential of an extra digit in the answer due to carry.

\paragraph{Parity.} For the parity task, we sample the length of each parity sequence from $1$ up to the maximum training length. We then sample randomly from $\{1,  0\}$ a sequence of the given length. We note that the definition of length we use is based on the sequence length and not based on the number of $1$s in the sequence. 

\paragraph{Boolean-AND.} For the boolean-AND task, we train on sequences of length $20$ and evaluate on test sequences of length $20$. To compare to the setting in \citet{abbe2023generalization}, we do not pack the context here and train on single examples in the context window. The training distribution has a $50\%$ chance of being a sequence of all $1$s, and a $50\%$ chance of having one $0$ element in the sequence. The position of the $0$ element is sampled uniformly between positions $0$ and $15$. The last $5$ elements in the training sequence are always $1$s. At test time, there is a $50\%$ chance of being a sequence of all $1$s, and a $50\%$ chance of having one $0$ element in the last $5$ elements in the sequence.

\begin{table*}[hbt!] 
\caption{Experimental hyperparameters. All experiments use AdamW optimizer and cosine learning rate schedule. Count, Copy, and Sort use weight decay of 0.1 and grad clip of 0. Parity and Mode use weight decay of 0.1 and grad clip of 1. Addition uses weight decay of 0 and grad clip of 1. Boolean-AND uses weight decay of 0 and grad clip of 0.}
\label{tab:exp_setting}
\begin{center}
\begin{tabular}{@{}llllllll@{}}
\toprule
\multicolumn{1}{c}{\small \bf Task}  &\multicolumn{1}{c}{\small \bf Model Size}
&\multicolumn{1}{c}{\small \bf Train Iter}&\multicolumn{1}{c}{\small \bf Context Len}&\multicolumn{1}{c}{\small \bf Batch Size}&\multicolumn{1}{c}{\small \bf Learning Rate} \\ \midrule
\small Count & 
\small $6$ layer; $8$ head; $64$ emb &
\small $10000$ &
\small $256$ &
\small $128$ &
\small 1e-3 to 1e-5 &
\\
\small Mode & \small $6$ layer; $8$ head; $512$ emb &
\small $10000$ &
\small $256$ &
\small $128$ &
\small 1e-3 to 1e-6 \\
\small Copy & \small $6$ layer; $8$ head; $512$ emb &
\small $100000$ &
\small $512$ &
\small $128$ &
\small 1e-4 to 1e-6 \\
\small Sort & \small $2$ layer; $16$ head; $1024$ emb &
\small $100000$ &
\small $1024$ &
\small $512$ &
\small 1e-5 to 0 \\
\small Addition & \small $6$ layer; $8$ head; $512$ emb &
\small $30000$ &
\small $512$ &
\small $64$ &
\small 1e-4 to 0 
\\
\small Parity & \small $6$ layer; $8$ head; $512$ emb &
\small $10000$ &
\small $512$ &
\small $256$ &
\small 1e-3 to 1e-6
\\
\small Boolean-AND & \small $2$ layer; $4$ head; $64$ emb &
\small $10000$ &
\small $128$ &
\small $128$ &
\small 1e-3 to 0
\\
\bottomrule
\end{tabular}
\end{center}
\end{table*}

\section{Appendix: Counterfactual Analysis on Count}
\label{app:counerfactual}

In this section, we probe whether models trained on count actually learn the count algorithm that we intuitively want. To reiterate, one simple algorithm that solves the count task is as follows. To predict the next token:

\hspace*{0.2cm}%
\begin{minipage}{.95\textwidth}
\begin{enumerate}
    \item Search for the most-recent \Verb|SoS| token, 
    and read the following two numbers as $a, b$.
    \item Read the previous token as $x$.
    If \verb|(x=='>')|, output $a$.
    If \verb|(x==b)|, output \Verb|EoS|.
    \item Otherwise, output $(x+1)$.
\end{enumerate}
\end{minipage}

\begin{figure}[t]
\captionsetup[subfigure]{justification=centering}
	\centering
	\subfloat[Counterfactual test of starting on first token]{\includegraphics[width=.5
	\linewidth]{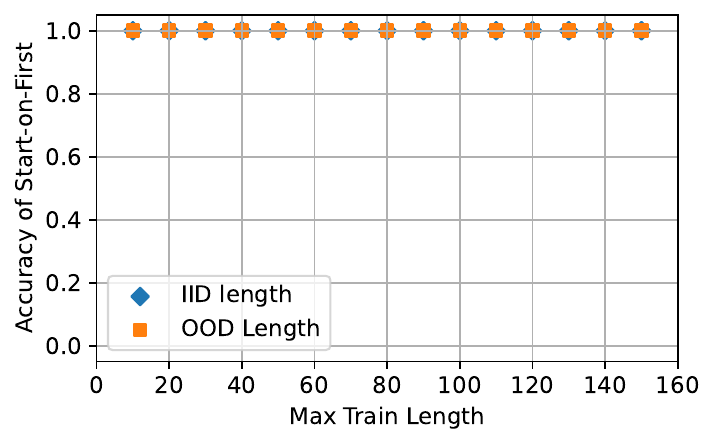}\label{fig:countfact-startfirst}}
	\subfloat[Counterfactual test of ending on last token]{\includegraphics[width=.5\linewidth]{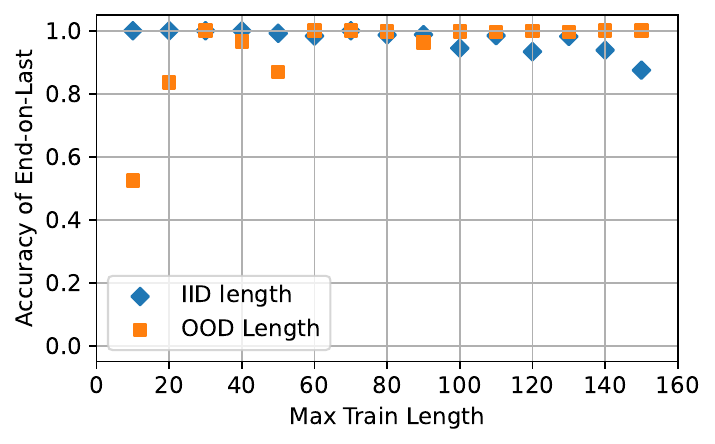}\label{fig:countfact-endonlast}}
	\caption{We measure performance of models trained on the count task on counterfactual tests designed to evaluate whether the model simulates the correct generalizing algorithm on random sequences far out-of-distribution. We see that models always output the start token as the first output, but do not always output the EoS token once the ending token has been outputted.
 }
	\label{fig:countfact-startend}
\end{figure}

\begin{figure}[t]
\captionsetup[subfigure]{justification=centering}
	\centering
	\subfloat[Counterfactual test of increment-by-1]{\includegraphics[width=.5
	\linewidth]{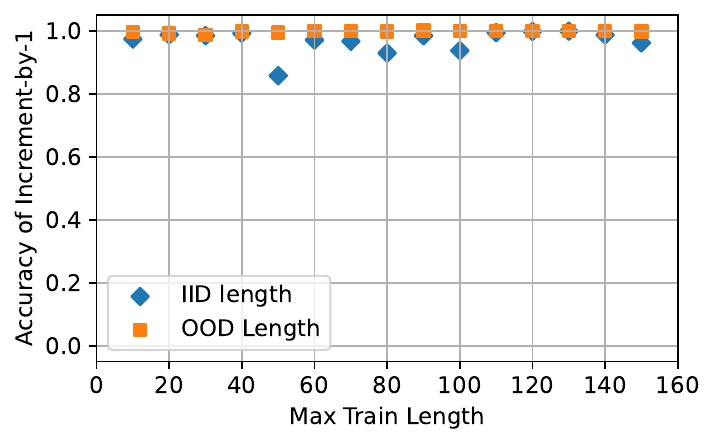}\label{fig:countfact-incby1}}
	\subfloat[Counterfactual test of the percentage of EoS outputs which are correct]{\includegraphics[width=.5\linewidth]{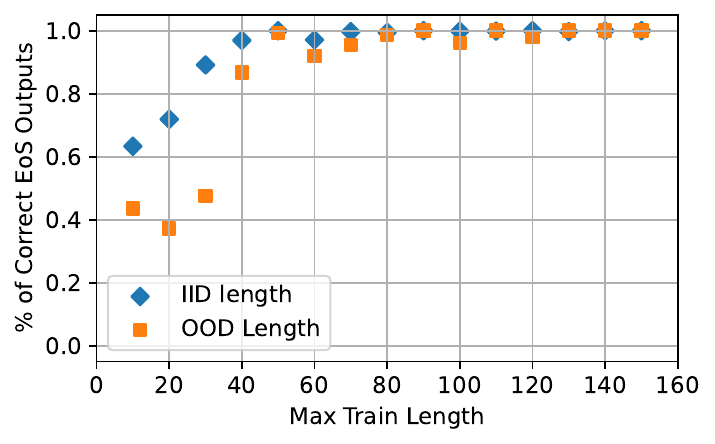}\label{fig:countfact-goodeos}}
	\caption{We measure performance of models trained on the count task on counterfactual tests designed to evaluate whether the model simulates the correct generalizing algorithm on random sequences far out-of-distribution. We see that models almost always increments the previous token by 1, no matter what the previous sequence is. However, it sometimes output the EoS token prematurely, especially on lengths longer than seen in training. This likely explains failures of length generalization observed in Figure~\ref{fig:count-lengen-scale}.
 }
	\label{fig:countfact-soph}
\end{figure}

Since there is no easy way to formally check if the Transformer model learns this exact algorithm internally, we employ the simple heuristic of running the model on counterfactual examples. The allows us to stress test the models behavior and see if it performs the expected algorithmic steps in an input-agnostic way. To do so, we performance inference on randomly generated input sequences that are designed to test the model in four ways:

\hspace*{0.2cm}%
\begin{minipage}{.95\textwidth}
\begin{enumerate}
    \item The model should always output the start token in the prompt ($a$) as the first output. (Figure~\ref{fig:countfact-startfirst})
    \item The model should always output \Verb|EoS| following a token that matches the ending token in the prompt ($b$). (Figure~\ref{fig:countfact-endonlast})
    \item In all other settings, the model should increment the previous token by $1$. (Figure~\ref{fig:countfact-incby1})
    \item The model should not output \Verb|EoS| prematurely. (Figure~\ref{fig:countfact-goodeos})
\end{enumerate}
\end{minipage}

We create the counterfactual dataset by sampling start and end tokens of varying distances, then generate a sequence of random tokens of the length specified by the distance between the start and end token. We then pass this sequence through a trained model and look at its predictions at each token position. The goal of the four proposed tests on random sequences is to probe whether the model learned the expected algorithmic behavior rather than learning something that would strongly depend on statistics of the training distribution. We sample examples for in-distribution lengths and out-of-distribution lengths based on the training distribution of each model. For simplicity, we choose $1$ model with strong length generalization performance from each maximum training length setting. The performance on each test is shown in Figure~\ref{fig:countfact-startend} and Figure~\ref{fig:countfact-soph}.

We see that for the start-on-first test and the increment-by-1 test, all models exhibit near perfect performance both in- and out-of-distribution. For the end-on-last test, we see that models trained with shorter lengths do not learn to robustly output \Verb|EoS| on long test sequences once the ending condition is met. However, on models trained on longer sequences (and has better length generalization), this behavior is more robust. Lastly, when we measure the percentage of \Verb|EoS| which are correct, we see that models that do not have strong generalization also fails to output \Verb|EoS| only at the appropriate time. This failing is observed on both in-distribution and out-of-distribution lengths. This suggests that the failures of length generalization can be attributed to prematurely outputting an \Verb|EoS| token before the full length of the sequence is outputted. Overall, we observe strong correspondence between the model's behavior and what we would expect from the correct algorithm. This lends credence to the intuition that the model learns the correct RASP-L program and generalizes because of it.

\section{Appendix: Additional Ablations}
\label{app:ablations}

In this section, we include some additional experiments to support the results in the main paper.

\begin{figure}[t]
\captionsetup[subfigure]{justification=centering}
	\centering
	\subfloat[Test EM on length 50 on sort task]{\includegraphics[width=.5
	\linewidth]{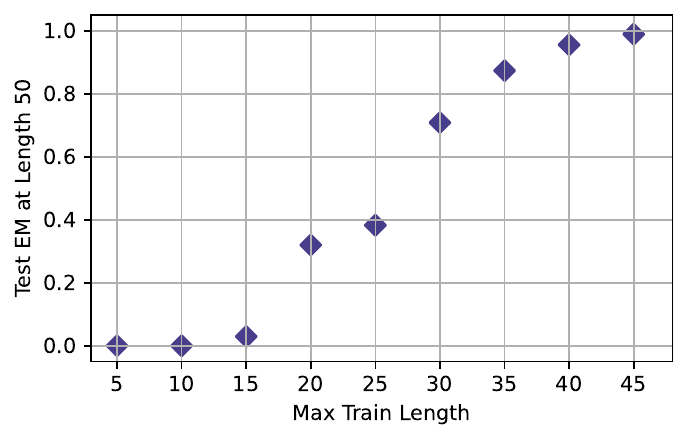}\label{fig:sort50}}
	\subfloat[Length generalization on sort task]{\includegraphics[width=.5\linewidth]{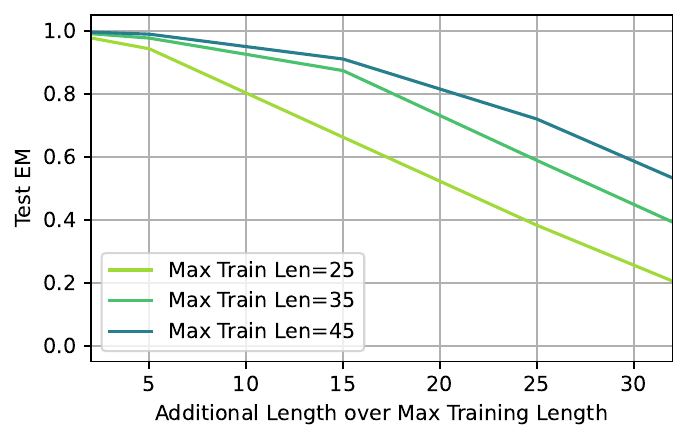}\label{fig:sortall}}
	\caption{Test EM for the sort task. We report the median performance over $20$ seeds. {\bf \protect\subref{fig:sort50}} measures test EM on length $50$ for models trained with various maximum training length. We see that models trained with $35$ or higher length show strong generalization to length $50$. {\bf \protect\subref{fig:sortall}} illustrates the generalization performance for select experiments as we go further out-of-distribution. We observe better generalization when models see greater coverage of sequence lengths at train time. 
 }
	\label{fig:sort}
\end{figure}

\subsection{Training Speed}
\label{app:trainspeed}
The RASP Generalization Conjecture suggests that simple-to-represent programs are also more likely to be learned. Intuitively, we should expect that simple-to-represent programs also exhibits faster optimization during training. 
In this section, we introduce a different scratchpad for the parity task, and see whether the training speed of these variants correspond to the simplicity of their corresponding RASP-L programs (and by extension their length generalization performance).

For the parity task, we introduce a more intuitive scratchpad format for comparison. This scratchpad outputs the sum-mod-10 of the parity sequence before outputting the final parity. An example is 
\texttt{
\setlength{\tabcolsep}{1pt}
\small
\begin{NiceTabular}{*{9}{c}}[hvlines]
0 & 0 & 1 & 1 & 0 & > & 2 & , & 0
\end{NiceTabular}}. This scratchpad does not simplify the problem as much as the previous one, because it does not leverage the autoregressive inference of the Transformer to process the task sequentially, and instead relying on a more complicated sum operation. However, it is still simpler than parity without any scratchpad because it helps to simplify the final operation of getting the parity of the sum. Instead of doing this internally, the model can now reference the output of the sum-mod-10 and learn a simple mapping between that and the corresponding parity. 
First, we compare the training speed for the different scratchpad formats.
Figure~\ref{fig:partrainspeed} shows the training curves for each parity format. We see that the main scratchpad (``Easy Scratchpad'') optimizes much more quickly than the sum-mod-10 scratchpad (``Hard Scratchpad''), even though the easy scratchpad has much longer output sequences. 
Next, we compare the length generalization performance for each scratchpad. We observe that Easy Scratchpad exhibits significantly stronger length generalization than Hard Scratchpad, shown in Figure~\ref{fig:parsumcot}. Both scratchpads optimizes much better than parity with no scratchpad, which is unable to even fit the training set and demonstrates no length generalization.

\subsection{Alternative scratchpad for mode}
\label{app:modescratch}

In Section~\ref{sec:scratchpad} we introduced a scratchpad for mode, which orders the intermediate counts in order of frequency. This may seem overly demanding, as it requires the model to know the order of the frequencies before outputting them. Another variant of this could output the scratchpad in order of appearance in the original sequence. Moreover, we can output the token first before outputting their count, which may help the model reference the relevant token for this step. An example is
\texttt{
\setlength{\tabcolsep}{3pt}
\small
\begin{NiceTabular}{*{16}{c}}[hvlines]
a & b & b & c & b & a & c & b & > & a & 2 & b & 4 & c & 2 & b
\end{NiceTabular}
}.

\begin{figure}[t]
\captionsetup[subfigure]{justification=centering}
	\centering
	\subfloat[Mode using scratchpad ordered by appearance]{\includegraphics[width=.5
	\linewidth]{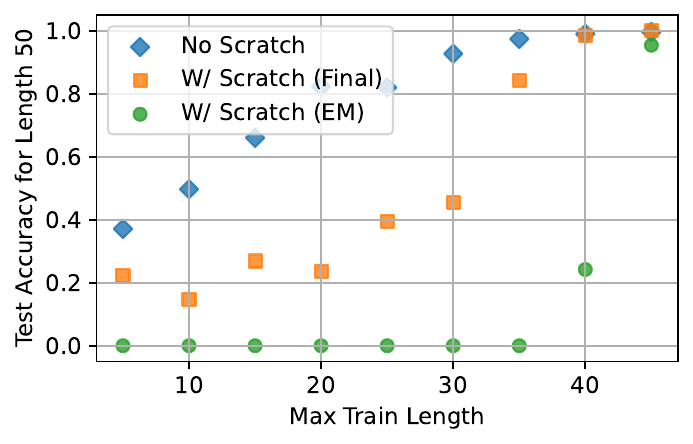}\label{fig:modeiapp}}
	\subfloat[Parity with sum-mod-10 scratchpad]{\includegraphics[width=.5\linewidth]{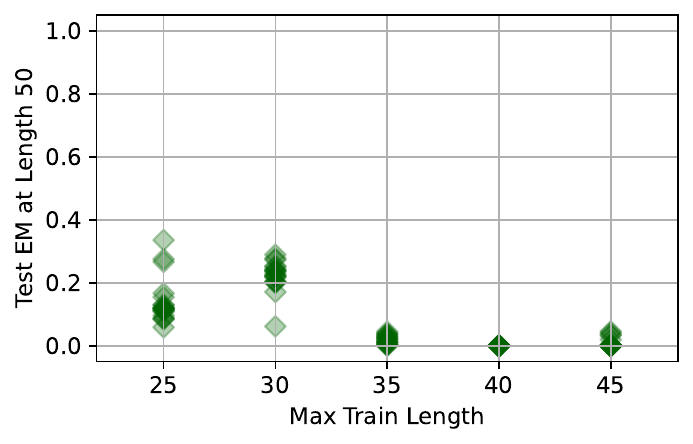}\label{fig:parsumcot}}
	\caption{{\bf \protect\subref{fig:modeiapp}} compares test performance of mode with or without scratchpad. In this case, we use the scratchpad presented in order of appearance. We see that no scratchpad significantly outperforms scratchpad, whether measured on the final answer accuracy or the exact match of the entire scratchpad output. {\bf \protect\subref{fig:parsumcot}} illustrates the generalization performance for parity with scratchpad on length $50$. We see that no runs show significant length generalization in this setting.
 }
	\label{fig:mode-inapp}
\end{figure}

The performance of this scratchpad is shown in Figure~\ref{fig:mode-inapp}. We see that utilizing this scratchpad still results in much worse length generalization performance than using no scratchpad.

\subsection{Performance with rotary embedding}
\label{app:rotary}
\begin{figure}[t]
\captionsetup[subfigure]{justification=centering}
	\centering
	\subfloat[Count with rotary embedding]{\includegraphics[width=.5
	\linewidth]{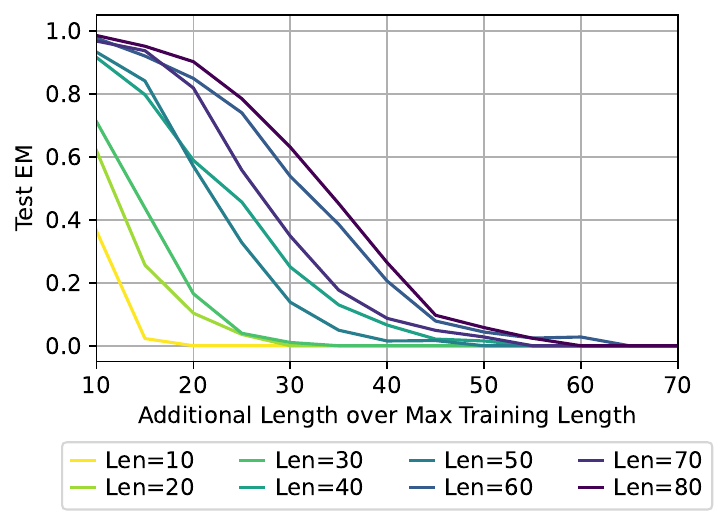}\label{fig:count-rotary}}
	\subfloat[Mode with rotary embedding]{\includegraphics[width=.5\linewidth]{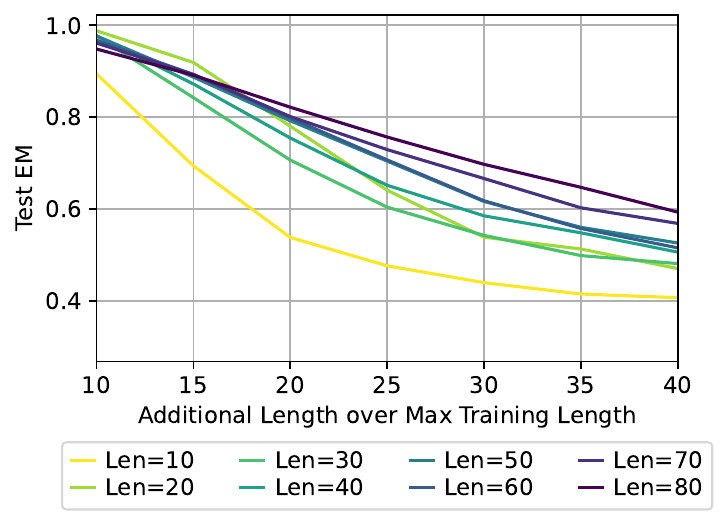}\label{fig:mode-rotary}}
	\caption{Length generalization performance on the count on mode tasks for models trained with rotary positional embedding.}
	\label{fig:count-mode-rotary}
\end{figure}

\begin{figure}[t]
\captionsetup[subfigure]{justification=centering}
	\centering
	\subfloat[Forward addition with rotary embedding]{\includegraphics[width=.5
	\linewidth]{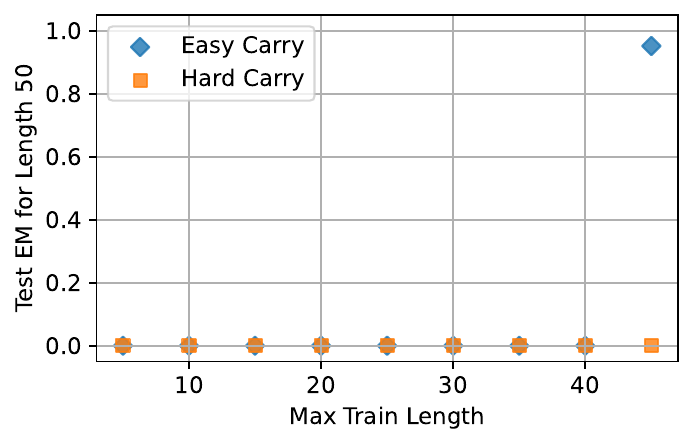}\label{fig:addfor-rotary}}
	\subfloat[Reverse addition with rotary embedding]{\includegraphics[width=.5\linewidth]{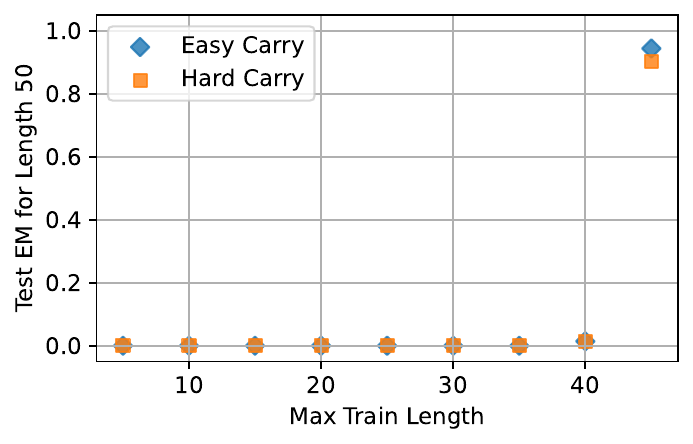}\label{fig:addrev-rotary}}
	\caption{Length generalization performance on the addition task with different scratchpads for models trained with rotary positional embedding.
 }
	\label{fig:add-rotary}
\end{figure}

\begin{figure}[t]
\captionsetup[subfigure]{justification=centering}
	\centering
	\subfloat[Parity with main scratchpad]{\includegraphics[width=.5
	\linewidth]{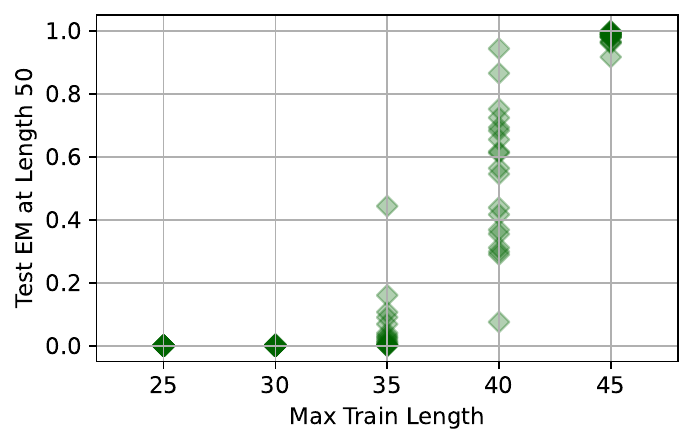}\label{fig:parletter-rotary}}
	\subfloat[Parity with sum-mod-10 scratchpad]{\includegraphics[width=.5\linewidth]{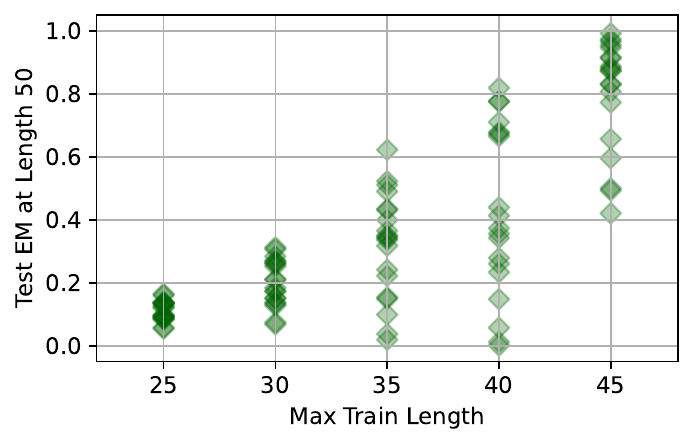}\label{fig:parsum-rotary}}
	\caption{Length generalization performance on the parity task with different scratchpads for models trained with rotary positional embedding.
 }
	\label{fig:par-rotary}
\end{figure}

For completeness, we also evaluate a number of tasks on models trained with rotary positional embedding. Result on count is shown in Figure~\ref{fig:count-rotary}. Result on mode is shown in Figure~\ref{fig:mode-rotary}. Results on addition with index hints are shown in Figure~\ref{fig:add-rotary}, and results for parity are shown in Figure~\ref{fig:par-rotary}. We observe worse length generalization across the tasks with rotary positional embedding. This is consistent with the findings reported in \citet{kazemnejad2023impact}, which showed that learned positional embedding is better than a number of more sophisticated embedding approaches on length generalization.

\begin{figure}[t]
\captionsetup[subfigure]{justification=centering}
	\centering
	\subfloat[Forward addition on easy carry examples]{\includegraphics[width=.5
	\linewidth]{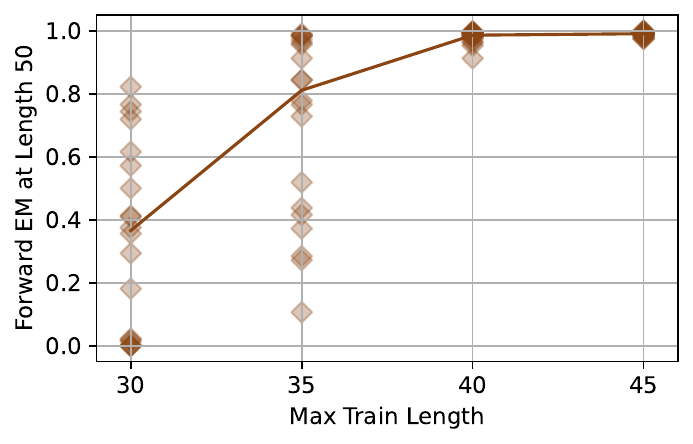}\label{fig:addforeasy}}
	\subfloat[Reverse addition on easy carry examples]{\includegraphics[width=.5\linewidth]{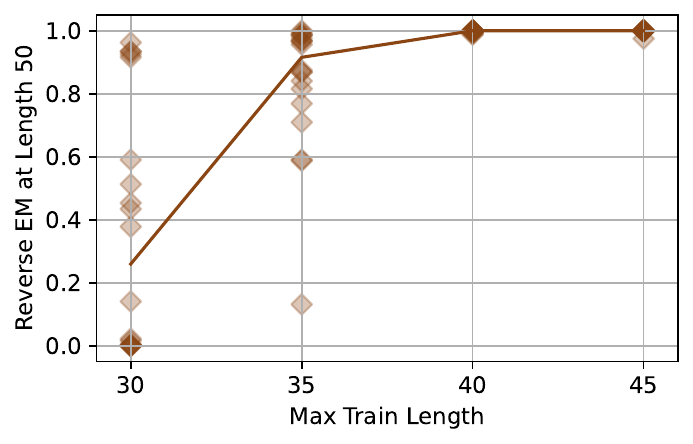}\label{fig:addreveasy}}
	\caption{Length generalization on addition with index hints. Each diamond shows the performance of one of $20$ runs, illustrating the spread of different training runs. {\bf \protect\subref{fig:addforeasy}} illustrates the generalization performance for forward addition with index hints on easy carry examples of length $50$. {\bf \protect\subref{fig:addreveasy}} illustrates the generalization performance for reverse addition with index hints on easy carry examples of length $50$. Easy carry examples consist of addition questions where the two numbers are sampled randomly and independently, which is the setting considered in prior works studying addition. We see that both settings demonstrate strong length generalization, thus demonstrating the usefulness of the index hints.
 }
	\label{fig:addeasy}
\end{figure}

\begin{figure}[t]
\captionsetup[subfigure]{justification=centering}
	\centering
	\subfloat[Forward addition trained with balanced carry]{\includegraphics[width=.5
	\linewidth]{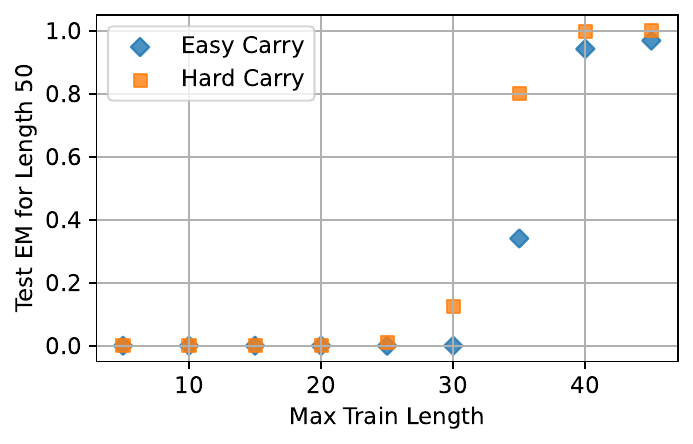}\label{fig:addforbalanced}}
	\subfloat[Reverse addition trained with balanced carry]{\includegraphics[width=.5\linewidth]{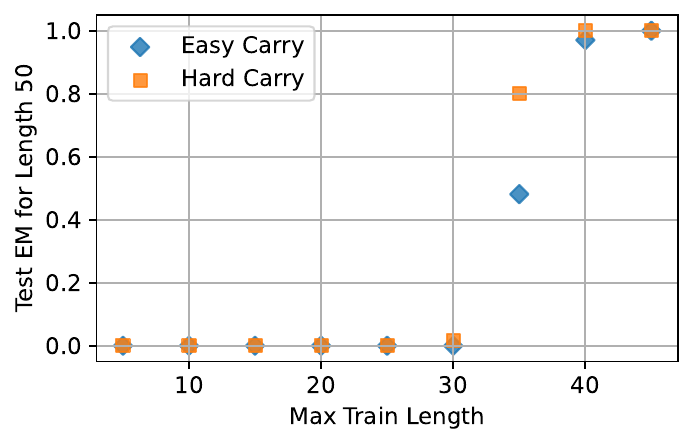}\label{fig:addrevbalanced}}
	\caption{Length generalization on addition with index hints, trained with balanced carry distribution. Each point shows the median test performance over $20$ runs. {\bf \protect\subref{fig:addforbalanced}} illustrates the generalization performance for forward addition with index hints on both easy and hard carry examples of length $50$. {\bf \protect\subref{fig:addrevbalanced}} illustrates the generalization performance for reverse addition with index hints on both easy and hard carry examples of length $50$. With balanced carry training, strong length generalization is observed for both forward and reverse addition on the hard carry evaluation setting. This demonstrates that with increased data diversity, we also increase the likelihood that the model learns a length-generalizing solution.
 }
	\label{fig:addbalanced}
\end{figure}

\clearpage
\input{RASP_L}

\section{RASP Programs}
\label{app:rasp_programs}
In this section, we provide the RASP-L core (\ref{rasp:core}) and library (\ref{rasp:library}); RASP-L programs for the tasks discussed in the paper, namely counting (\ref{rasp:count}), mode (\ref{rasp:mode}), copy-with-unique-tokens (\ref{rasp:copy_unique}), addition with forward and reverse order and index-hints (\ref{rasp:add_reverse_index_hints}); and a naive non-RASP addition algorithm (\ref{rasp:illegal_add}).

\begin{listing}[t]
\begin{minted}{python}
import numpy as np

def full(x, const):
    return np.full_like(x, const, dtype=int)
    
def indices(x):
    return np.arange(len(x), dtype=int)

def tok_map(x, func):
    return np.array([func(xi) for xi in x]).astype(int)

def seq_map(x , y, func):
    return np.array([func(xi, yi) for xi, yi in zip(x,y)]).astype(int)

def select(k, q, pred, causal=True):
    s = len(k)
    A = np.zeros((s, s), dtype=bool)
    for qi in range(s):
        for kj in (range(qi+1) if causal else range(s)):  # k_index <= q_index if causal
            A[qi, kj] = pred(k[kj], q[qi])
    return A

def sel_width(A):
    return np.dot(A, np.ones(len(A))).astype(int)

def aggr_mean(A, v, default=0):
    out = np.dot(A, v)
    norm = sel_width(A)
    out = np.divide(out, norm, out=np.full_like(v, default,dtype=float), where=(norm != 0))
    return out.astype(int)

def aggr_max(A, v, default=0):
    out = np.full_like(v, default)
    for i, row in enumerate(A):
        idxs = np.flatnonzero(row)
        if len(idxs) > 0:
            out[i] = np.max(v[idxs]) # max of selected elements in v
    return out.astype(int)

def aggr(A, v, default=0, reduction='mean'):
    if reduction == 'mean':
        return aggr_mean(A, v, default)
    elif reduction == 'max':
        return aggr_max(A, v, default)
    elif reduction == 'min':
        return -aggr_max(A, -v, -default)

def kqv(k, q, v, pred, default=0, reduction='mean'):
    return aggr(select(k, q, pred), v, default=default, reduction=reduction)
\end{minted}
\caption{RASP-L core functions.}
\label{raspL-core}
\end{listing}

\begin{listing}[t]
\begin{minted}{python}
import operator as op
equals, leq, lt, geq, gt = op.eq, op.le, op.lt, op.ge, op.gt

def shift_right(x, n, default=0):
    # shifts sequence x to the right by n positions
    return kqv(indices(x) + n, indices(x), x, equals, default=default)

def cumsum(bool_array):
    # returns number of previous True elements in bool_array
    return sel_width(select(bool_array, bool_array, lambda k, q: k))

def where(condition, x_if, y_else):
    # equivalent to np.where(condition, x_if, y_else)
    x_masked = seq_map(x_if,    condition, lambda x, m: x if m else 0)
    y_masked = seq_map(y_else,  condition, lambda y, m: y if not m else 0)
    return seq_map(x_masked, y_masked, lambda x, y: x if y == 0 else y)

def mask(x, bool_mask, mask_val=0):
    # equivalent to x*bool_mask + default*(~bool_mask)
    return where(bool_mask, x, full(x, mask_val))

def maximum(x):
    return kqv(x, x, x, lambda k, q: True, reduction='max')

def minimum(x):
    return -maximum(-x)

def argmax(x):
    mm = maximum(x)
    return kqv(mm, x, indices(x), reduction='max')

def argmin(x):
    return argmax(-x)

def num_prev(x, queries):
    # output[i] = number of previous elements of x equal to queries[i], inclusive
    return sel_width(select(x, queries, equals))

def has_seen(x, queries):
    return kqv(x, queries, full(x, 1), equals, default=0)

def firsts(x, queries, default=-1):
    # find the index of the first occurrence of each query[i] in x
    # out[i] := np.flatnonzero(x[:i+1] == queries[i]).min() 
    return kqv(x, queries, indices(x), equals, default=default, reduction='min')

def lasts(x, queries, default=-1):
    # find the index of the last occurrence of each query[i] in x
    # out[i] := np.flatnonzero(x[:i+1] == queries[i]).max() 
    return kqv(x, queries, indices(x), equals, default=default, reduction='max')

def index_select(x, idx, default=0):
    # indexes into sequence x, via index sequence idx
    # i.e. return x[idx] if idx[i] <= i else default
    return kqv(indices(x), idx, x, equals, default=default)

def first_true(x, default=-1):
    # returns the index of the first true value in x
    seen_true = kqv(x, full(x, 1), full(x, 1), equals, default=0)
    first_occ = seq_map(seen_true, shift_right(seen_true, 1), lambda curr, prev : curr and not prev)
    return kqv(first_occ, full(x, 1), indices(x), equals, default=default)

def induct_kqv(k, q, v, offset, default=0, null_val=-999):
    # get value of v at index of: first occurrence of q[i] found in k (if found) + offset.
    # (excludes the last OFFSET tokens of k from matching)
    # null_val is a special token that cannot appear in k or q; used to prevent accidental matches
    indices_to_copy = firsts(shift_right(k, offset, default=null_val), q, default=null_val)
    copied_values = index_select(v, indices_to_copy, default=default)
    return copied_values

def induct(k, q, offset, default=0, null_val=-999):
    return induct_kqv(k, q, k, offset=offset, default=default, null_val=null_val)

def induct_prev(k, q, offset, default=0, null_val=-999):
    # A version of induct for negative offsets.
    indices_to_copy = firsts(k, q, default=null_val) + offset
    copied_values = index_select(k, indices_to_copy, default=default)
    return copied_values
\end{minted}
\caption{RASP-L library functions.}
\label{rasp:library}
\end{listing}

\begin{listing}[t]
\begin{minted}{python}
def count(seq):
    # First, find the index of most-recent START_TOK (begninng of current sequence)
    start_idx = find_last_tok(seq, START_TOK)
    
    # Then, compute the start/end numbers of the current sequence
    start_nums = index_select(seq, start_idx+1)
    end_nums   = index_select(seq, start_idx+2)

    # Bool arrays: whether we're predicting the first / last tokens of the current sequence
    pred_first_pos = seq_map(indices(seq), start_idx+2, equals)
    pred_final_pos = (~pred_first_pos) & seq_map(seq, end_nums, equals) 

    next_tok = where(pred_first_pos,             # if predicting the first token:
                     start_nums,                 #    next_tok = starting num
                     where(pred_final_pos,       # else if predicting the final token:
                           full(seq, END_TOK),   #    next_tok = END_TOK                
                           seq + 1))             # else: next_tok = prev_tok + 1
    return next_tok
\end{minted}
\caption{RASP-L program for Count.}
\label{rasp:count}
\end{listing}

\begin{listing}[t]
\begin{minted}{python}
def mode(x):
    num_prev_matching = sel_width(select(x, x, equals))
    idx = argmax(num_prev_matching)
    return index_select(x, idx)
    
def binary_mode(x):
    num_prev_zeros = sel_width(select(x, full(x, 0), equals))
    num_prev_ones  = sel_width(select(x, full(x, 1), equals))
    mode_val = seq_map(num_prev_ones, num_prev_zeros, gt)*1
    return mode_val
\end{minted}
\caption{RASP-L program for Mode.}
\label{rasp:mode}
\end{listing}

\begin{listing}[t]
\begin{minted}{python}
def copy_unique(seq):
    return induct(seq, seq, offset=1)

def copy_unique_ar(x):
    START, END = -1, -2
    prompt = np.concatenate(([START], x, [END], [START]))
    seq = prompt.copy()

    while seq[-1] != END:
        next_tok = copy_unique(seq)[-1]
        seq = np.concatenate((seq, [next_tok]))
    return seq

copy_unique_ar(np.array([8, 3, 4, 2, 1, 5]))
>> [-1  8  3  4  2  1  5 -2 -1  8  3  4  2  1  5 -2]
\end{minted}
\caption{RASP-L program for Copy with unique tokens.}
\label{rasp:copy_unique}
\end{listing}

\begin{listing}[t]
\begin{minted}[fontsize=\tiny,frame=single,]{python}
def next_tok_sort(x): # to be used auto-regressivly
    return kqv(x, x, x, gt, reduction='min') # min token in context that is > current token
\end{minted}
\caption{RASP-L program for Sort.}
\label{rasp:sort}
\end{listing}

\begin{listing}[t]
\begin{minted}{python}
def add_illegal(inp):  # inp = array of zero-padded digits of x0, x1; returns z = x0 + x1
    num_digits = int(len(inp)/2)  # ILLEGAL: no division on index types
    z = np.zeros(num_digits)
    carry = 0
    reversed_range = range(num_digits)[::-1]  # ILLEGAL: reversal is non-causal
    for i in reversed_range:  # ILLEGAL: no for loops
        x0, x1 = inp[i], inp[num_digits+i]  # ILLEGAL: variables cannot be used as indices
        digit_sum = x0 + x1
        z[i] = (digit_sum + carry) % 10
        carry = 1 if digit_sum > 9 else 0
    return z
\end{minted}
\caption{An addition program that works in Python but is illegal in RASP-L for several reasons.}
\label{rasp:illegal_add}
\end{listing}

\input{app_rasp_addition}

\input{rasp_aggr}

\section{Technical Lemma}

Lemma~\ref{lem:helper} follows directly from the following fact.
We state this with the boolean hypercube identified with $\{\pm 1\}^n$ 
as is standard in boolean function analysis, but these statements can be translated to $\{0, 1\}^n$.

\begin{lemma}
\label{lem:helper}
Let $f: \{\pm 1\}^n \to \R$
be a boolean function.
Suppose $f$ depends on its first coordinate; that is
suppose $\exists z \in \{\pm 1\}^{n-1}: ~~f(1 \circ z) \neq f(-1 \circ z)$.
Then, restricting the first coordinate of $f$ to be $1$ strictly reduces its degree profile:
\[
\mathrm{DegP}(f\vert_{x_1 = 1}) < \mathrm{DegP}(f)
\]
\end{lemma}
\begin{proof}
(Sketch).
Consider the multilinear representation of $f$, and factor out terms containing $x_1$:
\begin{align}
f(x_1, \bar{x}) &= x_1 P(\bar{x}) + Q(\bar{x}) 
\end{align}
where $\bar{x}$ denotes $(x_2, x_3, \dots, x_n)$.
When restricted with $x_1=1$, we have
\begin{align}
f(x_1, \bar{x})\vert_{x_1 = 1} &= P(\bar{x}) + Q(\bar{x}) 
\end{align}
Now, because $f$ depends on $x_1$ by assumption, we must have $P \neq 0$.
Thus, comparing the above two, at least one monomial containing $x_1$
has reduced degree with the restriction. The conclusion follows.
\end{proof}

%% file: RASP_L.tex
\section{RASP Specification}
\label{sec:rasp_spec}

Here we describe RASP-L as a restricted subset of Python (with numpy).
First, every RASP-L program accepts an input sequence of length $n$, for arbitrary $n \in \N$, and
returns an output sequence of the exact same length--- just like a transformer.
The restrictions on Python are:
All variables are either numpy arrays of length $n$ (``sequences''), or binary matrices in $\{0, 1\}^{n \x n}$ (``selectors'').
No control flow is allowed; all programs are straight-line programs, with no branching or loops.
Finally, every line of the program must be a call to either one of the core functions defined in
Listing~\ref{raspL-core}, or to another RASP-L program.

The original RASP technically allows for
certain operations which are possible to represent,
but not ``easy'' to represent or learn.
For example, any arbitrarily-complex tokenwise operation
$\R \to \R$ 
is allowed.
To disallow such pathologies,
RASP-L enforces the following additional restrictions.

First, all non-index values, including the input and
intermediate variables,
must be of type \bint.
This constraint handles issues with both infinite-precision
and with learnability of the tokenwise operations
(since all tokenwise functions now have small finite domains and co-domains, $\bint \to \bint$,
and so can be easily memorized).
Although disallowing floating-point operations and unbounded numbers
may seem like a strong restriction, we find these are not necessary
for most symbolic programs.

Moreover, token indices are treated specially.
In RASP-L we only allow the following operations on indices:
order comparisons with other indices,
and computing successor/predecessor.
This can be formalized via a type system:
the RASP-L core function \texttt{indices(x)}
formally returns a special type \indexint, which
can take values in $\N$, but only allows these restricted operations.
That is, we allow adding 1 to an \indexint,
but we do not allow adding two \indexint{s}, nor casting between \indexint~ and \bint.

There is one additional restriction, involving the ``selector width'' core operation
(\verb|sel_width|).
Selector-width returns the number of prior elements
that are selected by a binary Attention matrix,
for each token position.
The return type of Selector-width in RASP-L
inherits from \indexint: thus it can represent unbounded numbers of selected elements,
but can only operate on them in restricted ways.
Moreover, every call to \verb|sel_width| in RASP-L returns a \emph{new type}
which inherits from \indexint.
No operations are defined between distinct types --
that is, the sequences returned by two different calls to \verb|sel_width| are incomparable.

The reason for these restrictions, which may otherwise seem contrived, is that
\verb|sel_width| can be used to simulate \verb|indices|, by calling it on the all-ones selector matrix.
Thus, we must restrict the output of \verb|sel_width| sufficiently to not allow bypassing
the intended restrictions on index-arithmetic.
There may also be more mechanistic motivations for such restrictions, 
since the Transformer implementation of selector width requires
weights which grow linearly with sequence length \citep{lindner2023tracr}.
Finally, we acknowledge that these proposed language restrictions may not be the best way
of capturing the intended index restrictions,
but we consider it a reasonable and usable candidate.
Fully formalizing the intended index restrictions, 
and proving they hold under the appropriate type system,
is an open question for future work.

%% file: app_rasp_addition.tex
\subsection{Forward and Reverse Addition in RASP-L}
\label{sec:raspadd}

We provide a RASP-L program for addition (with output reversed, and index-hints)
in Listing~\ref{rasp:add_reverse_index_hints}.
This program specifies the next-token function; to compute the entire
output sequence, it must be called autoregressively (e.g. as in Listing~\ref{rasp:add_infra}).

The expected format of the prompt is:
\begin{center}
{\small
\verb|[START_PROMPT, <first summand>, PLUS, <second summand>, EQUALS_SIGN]|
}.
\end{center}
Both summands are given with index-hints; 
these hints start at $-100$ and decrement by $1$ for each
decimal position.
Further, both summands must have the same number of digits (the shorter summand is padded with $0$s if neccesary), and both must start with a $0$.
For example,
the prompt for ``$88 + 842 = 930$''
is encoded as:
\[
\textrm{[-1, -100, 0, -101, 0, -102, 8, -103, 8, -2, -100, 0, -101, 8, -102, 4, -103, 2, -3]}
\]
where we use the constants defined at the top of Listing~\ref{rasp:add_reverse_index_hints}
(e.g. \texttt{START\_PROMPT}$=-1$, \texttt{PLUS}$=-2$, \texttt{EQUALS\_SIGN}$=-3$).
For addition in reverse order, the expected autoregressivly-generated full sequence is:
\begin{center}
\small
[-1, -100, 0, -101, 0, -102, 8, -103, 8, -2, -100, 0, -101, 8, -102, 4, -103, 2, -3, -103, 0, -102, 3, -101, 9, -100, 0, -5].
\end{center}

For addition with output in \emph{standard} order,
we replace the highlighted codeblock in Listing~\ref{rasp:add_reverse_index_hints}
with the patch in Listing~\ref{rasp:fwd-block}.
The expected entire output sequence in this case is:
\begin{center}
\small
[-1, -100, 0, -101, 0, -102, 8, -103, 8, -2, -100, 0, -101, 8, -102, 4, -103, 2, -3, -100, 0, -101, 9, -102, 3, -103, 0, -5].
\end{center}

\begin{listing}[p]
\begin{minted}[linenos,fontsize=\tiny,frame=single,highlightlines={33-43}]{python}
## Constants and helper functions
START_PROMPT = -1
PLUS = -2
EQUALS_SIGN = -3
END_RESPONSE = -5
NONE = -88

def mask_between_tokens(seq, tok0, tok1):
    seen_tok0 = has_seen(seq, full(seq, tok0))
    seen_tok1 = has_seen(seq, full(seq, tok1))  
    ind_between = seq_map(seen_tok0, seen_tok1, lambda a, b: a and not b)  # ind(tok0) <= (*) < ind(tok1)
    return ind_between

def _add_safe(x, y):
    return x + y if (x >= 0) else x # preserve index-hints

## Next-token function
def next_tok_rev_addition_hinted(seq):
    prompt_mask = 1-has_seen(seq, full(seq, EQUALS_SIGN))
    second_summand_mask = mask_between_tokens(seq, PLUS, EQUALS_SIGN)
    prompt = mask(seq, prompt_mask)
    
    # let's first align the 1st summand with the second.
    other_summand_digit = induct(k=prompt, q=shift_right(prompt, 1), offset=1)
    pairsums = seq_map(seq, other_summand_digit, _add_safe)  # this aligns pairsums with the 2nd summand
    pairsums = mask(pairsums, second_summand_mask, NONE)
    pairsums_nh = mask(pairsums, (seq >= 0), NONE) # no hints: only keep digits
    
    curr_output_digit  = shift_right(seq, 1)
    curr_pairsum = induct(pairsums, shift_right(seq, 2), offset=1) # pairsum that generated curr_output_digit
    next_pairsum = induct(pairsums, seq, offset=1)

    ## START CHANGES
    direct_carry = curr_pairsum > 9  # previous sum gives carry
    indirect_carry = (curr_pairsum == 9) & (curr_output_digit == 0)  # previous sum is 9 and earlier sum gave carry
    next_tok_gets_carry = direct_carry | indirect_carry

    # (simple) index-hint computations:
    final_hint = full(seq, -100) # final hint output is always -100 
    first_hint =  induct_prev(seq, full(seq, EQUALS_SIGN), offset=-2) # first hint is 2 places before '=' 
    next_hint = shift_right(seq, 1) + 1 
    eos = (next_hint > final_hint)
    ## END CHANGES

    next_tok = next_pairsum
    next_tok += next_tok_gets_carry
    next_tok = next_tok %

    ## Finally, handle the case of outputing index-hints
    next_tok_is_index_hint = (seq > -100) # all index-hints are <= -100
    eos = (eos & next_tok_is_index_hint)

    next_tok = where( next_tok_is_index_hint, next_hint, next_tok)
    next_tok = where( eos, full(seq, END_RESPONSE), next_tok)
    next_tok = where( (seq == EQUALS_SIGN), first_hint, next_tok) 
    return next_tok
\end{minted}
\caption{RASP-L program for addition, with output in reverse order, and index-hints.
See Section~\ref{sec:raspadd} for details on prompt format.
For addition in forward order, the highlighted codeblock is replaced with Listing~\ref{rasp:fwd-block}.
}
\label{rasp:add_reverse_index_hints}
\end{listing}
\begin{listing}[p]
\begin{minted}[linenos,fontsize=\tiny,frame=single,highlightlines={33-43}]{python}
    ## START CHANGES
    gives_carry = tok_map(pairsums_nh, lambda _x: 1 if _x > 9 else 0)
    z = cumsum((pairsums_nh != 9) & (pairsums_nh != NONE))
    u = mask(z, gives_carry, mask_val=NONE)
    v = tok_map(u, lambda _x: _x - 1)
    chain_end_idxs = firsts(z, v, default=NONE)   # (left) ending indices of carry-chain
    
    curr_tok_got_carry = ((curr_pairsum %
    next_tok_inside_carry_chain =  (next_pairsum == 9) & curr_tok_got_carry 
        # in the middle of a carry-chain? (NOTE: assumes the pairsums has first element 0)
      
    next_tok_idx = kqv(pairsums, seq, indices(seq), equals) + 1
        # which answer-position are we at? (indices aligned to pairsums)
    next_tok_chain_end = kqv( chain_end_idxs , next_tok_idx , full(seq, 1), equals, default=0)
        # does the next_tok get a carry from the end of a carry-chain?
    next_tok_gets_carry = next_tok_inside_carry_chain | next_tok_chain_end

    # (simple) index-hint computations:
    final_hint = induct_prev(seq, full(seq, EQUALS_SIGN), offset=-2) # final hint is 2 places before '='
    first_hint = full(seq, -100)  
    next_hint = shift_right(seq, 1) - 1 
    eos = (next_hint < final_hint)
    ## END CHANGES
\end{minted}
\caption{The required patch to Listing~\ref{rasp:add_reverse_index_hints}, to 
produce a program for addition in forward order.
This code replaces the highlighted block in Listing~\ref{rasp:add_reverse_index_hints}.
See Section~\ref{sec:raspadd} for details on prompt format.
}
\label{rasp:fwd-block}
\end{listing}

\begin{listing}[t]
\begin{minted}[fontsize=\tiny,frame=single,]{python}
def sample_autoregressive(prompt, func=next_tok_rev_addition_hinted):
    seq = prompt.copy()
    while seq[-1] != END_RESPONSE:
        next_tok = func(seq)[-1]
        seq = np.concatenate((seq, [next_tok]))
    return seq
\end{minted}
\caption{Simulating autoregressive sampling given a RASP-L next-token function.}
\label{rasp:add_infra}
\end{listing}

%% file: rasp_aggr.tex
\clearpage
\section{RASP-L Aggregations}
\label{sec:rasp-aggr}

Here we describe how the ``max'' and ``min'' aggregations of RASP-L
can be represented by an Attention layer.
First, we review the standard construction for mean-aggregation (e.g. \citet{lindner2023tracr}).

\subsection{Standard Mean-Aggregation}
The standard Select-Aggregate in the original RASP is defined as follows.
We are given sequences $q, k, v$ and predicate $P: \Vocab \x \Vocab \to \{0, 1\}$.

We want to compute:\\
\verb|out = kqv(k,q,v, P) := aggr(select(q, k, P), v)| \\
which is equivalent to:
\[
\mathrm{out} = S \cdot v
\]
where
\begin{align*}
S_{i, j} &:= \frac{M_{ij}}{\sum_j M_{ij}}
\quad ; \quad
M_{i, j} := P(q_i, k_j).
\end{align*}
Note, the attention matrix $S$ is the row-normalized version of $M$.
Rows of $S$ are indexed by queries, and columns indexed by keys.

This is compiled into a Transformer as follows.
Given sequences $q, k \in \Vocab^T$ in a finite alphabet $\Vocab$,
and predicate $P$, we can construct vectors
$Q \in \{0, 1\}^{T \x |\Vocab|}$
and $K \in \{0, 1\}^{T \x |\Vocab|}$ such that
\[
\langle Q_{i, \cdot}, K_{j, \cdot} \rangle
=
P(q_i, k_j) \in \{0, 1\}.
\]
Explicitly, this is by one-hot encoding:
\begin{align}
K_{j, m} &= 
P(\Vocab[m], k_j)\\
Q_{i, n} &=
\1\{ q_i = \Vocab[n] \}
\end{align}
where $\Vocab[m]$ denotes the $m$-th element of the alphabet $\Vocab$.

Using these $Q, K$ sequences as input to an Attention Layer
yields the following ``pre-softmax attention matrix'' $M$:
\begin{align}
\label{eqn:boolattn}    
M_{i, j} = P(q_i, k_j).
\end{align}
Now, let $\sigma(\cdot)$ denote the row-wise softmax operator, at 0-temperature.
Applying this yields the final attention matrix:
\[
S_{i, j} = \sigma(M)_{i, j} =
\frac{M_{ij}}{\sum_j M_{ij}}.
\]
as desired.

\subsection{Max-Aggregation}
Now let us adapt the same construction for max-aggregation.
We want to implement an operation:
$\mathrm{kqv}_{\textrm{max}}(k, q, v, P)$
which has output $S \cdot v$, where
the attention matrix
\[
S_{i,j} = \1\{j = \argmax_m v_m P(q_i, k_m) \}
\]

This can be compiled into a Transformer as follows.
We use the following simple lemma, which is trivial to prove.
\begin{lemma}[Constructability] 
We call a pre-softmax matrix $M$ ``constructable'' if there exist
transformer weights which produce $M$ at the attention-layer (pre-softmax).
Now, if two matrices $M_1, M_2$ are constructable,
then so is any linear combination $\alpha M_1 + \beta M_2$ for $\alpha, \beta \in \R$.
\end{lemma}

Now, consider the matrix
\[
Z_{i, j} = v_j \in \{1, 2, 3, \dots, |\Vocab|\}.
\]
where we have identified the alphabet $\Vocab$ with natural numbers.
The matrix $Z$ is clearly constructable in the sense of the Lemma above.

Let $M$ be the matrix from Equation~\ref{eqn:boolattn} above, in the standard construction of mean-aggregation.
Now $M$ and $Z$ are both constructable, so the matrix 
\[
M^* = Z +2|\Vocab|M_1
\]
is also constructable, by the lemma.
When we apply the softmax to this, we get the attention matrix
\begin{align}
S_{i, j}
&= \sigma(M^*)_{i, j}\\
&= \1\{j = \argmax_m (v_m + 2|\Vocab|P(q_i, k_m)) \} \\
&= \1\{j = \argmax_{m: P(q_i, k_m) = 1} v_m \} \tag{since $2|\Vocab| > v_m$ for all $m$}\\
&= \1\{j = \argmax_m v_m P(q_i, k_m) \}
\end{align}
as desired.